\documentclass[journal]{IEEEtai}

\usepackage[colorlinks,urlcolor=blue,linkcolor=blue,citecolor=blue]{hyperref}

\usepackage{color,array}
\usepackage{caption}
\usepackage{multirow}%
\usepackage{amsmath,amssymb,amsfonts}%
\usepackage{amsthm}%
\usepackage{mathrsfs}%
\usepackage{xcolor}%
\usepackage{textcomp}%
\usepackage{manyfoot}%
\usepackage{booktabs}%
\usepackage{algorithm}%
\usepackage{algorithmicx}%
\usepackage{algpseudocode}%
\usepackage{listings}

\usepackage{graphicx}
\usepackage{tabularx}
\usepackage{subcaption}
\usepackage{tikz}
\usetikzlibrary{arrows.meta,positioning,calc,fit,backgrounds,shapes.geometric}
\tikzset{
  flowbox/.style={draw=black!70, rounded corners=2pt, fill=black!4, align=center, font=\footnotesize, minimum height=9mm, inner sep=3pt},
  autobox/.style={draw=black!70, rounded corners=2pt, fill=blue!8, align=center, font=\footnotesize, minimum height=9mm, inner sep=3pt},
  flowarr/.style={-{Stealth[length=2mm]}, thick},
  autoarr/.style={-{Stealth[length=2mm]}, thick, dashed, draw=black!60},
  treecat/.style={draw=black!70, rounded corners=2pt, fill=black!6, align=left, font=\scriptsize\bfseries, inner sep=3pt},
  treeleaf/.style={draw=black!50, rounded corners=1pt, fill=white, align=left, font=\scriptsize, inner sep=3pt},
}
\newcolumntype{Y}{>{\raggedright\arraybackslash}X}
\definecolor{cMDP}{rgb}{0.10,0.50,0.52}
\definecolor{cALG}{rgb}{0.80,0.47,0.12}
\definecolor{cHPO}{rgb}{0.72,0.22,0.26}
\definecolor{cL2L}{rgb}{0.44,0.34,0.63}
\definecolor{cNAS}{rgb}{0.22,0.42,0.68}
\definecolor{cLLM}{rgb}{0.27,0.52,0.28}

\begin{document}

\title{Automated Reinforcement Learning: An Overview}

\author{Reza Refaei Afshar, Joaquin Vanschoren, Uzay Kaymak, Rui Zhang, Yaoxin Wu, Wen Song, Yingqian Zhang
\thanks{Reza Refaei Afshar, Joaquin Vanschoren, Yaoxin Wu and  Yingqian Zhang are with Eindhoven University of Technology, 5600 MB Eindhoven, the Netherlands (e-mails: r.refaei.afshar@tue.nl,  j.vanschoren@tue.nl, y.wu2@tue.nl, yqzhang@tue.nl).}
\thanks{Uzay Kaymak, is with Jheronimus Academy of Data Science (JADS), `s-Hertogenbosch, the Netherlands (e-mail: u.kaymak@ieee.org).}
\thanks{Rui Zhang and Wen Song are with Shandong University, 266237 Qingdao, China (e-mails: 202300190245@mail.sdu.edu.cn, wensong@email.sdu.edu.cn).}}

% \markboth{Journal of IEEE Transactions on Artificial Intelligence, Vol. 00, No. 0, September 2024}
% {Reza Refaei Afshar \MakeLowercase{\textit{et al.}}}

\maketitle

\begin{abstract}
Reinforcement Learning and, recently, Deep Reinforcement Learning are popular methods for solving sequential decision-making problems modeled as Markov Decision Processes. RL modeling of a problem and selecting algorithms and hyper-parameters require careful consideration, as different configurations may entail completely different performances. These considerations are mainly the task of RL experts; however, RL is progressively becoming popular in other fields, such as combinatorial optimization, where researchers and system designers are not necessarily  RL experts. Besides, many modeling decisions are typically made manually, such as defining state and action space, size of batches, batch update frequency, and timesteps. For these reasons, automating different components of RL is of great importance, and it has attracted much attention in recent years. Automated RL provides a framework in which different components of RL, including MDP modeling, algorithm selection, and hyper-parameter optimization, are modeled and defined automatically. In this article, we present the literature on automated RL (AutoRL), including the recent large language model (LLM) based techniques. We also discuss the recent work on techniques that are not presently tailored for automated RL but hold promise for future integration into AutoRL. 
Furthermore, we discuss the challenges, open questions, and research directions in AutoRL.
\end{abstract}

\begin{IEEEImpStatement}
Automated Reinforcement Learning (AutoRL) aims to make reinforcement learning accessible to non-experts by automating complex processes such as MDP modeling, algorithm selection, and hyper-parameter optimization. This paper provides a comprehensive overview of AutoRL, exploring its potential to significantly impact fields like robotics, optimization, and control systems by reducing the need for extensive RL expertise. The advancements discussed in the article will enable broader adoption and application of RL methods, fostering innovation and efficiency in various domains of artificial intelligence and complex decision-making.
\end{IEEEImpStatement}


\section{Introduction}
\label{Introduction}
\IEEEPARstart{R}{einforcement} Learning (RL) is a learning approach in which no prior knowledge of an environment is necessary. In this definition, the environment is any factor surrounding an agent that provides states and applies action signals. An agent learns the optimal behavior known as \textit{policy} by interacting with the environment. At each decision step, the agent observes the \textit{state} of the environment, takes an \textit{action}, and receives a scalar \textit{reward} value from the environment. Using this reward value, the agent adjusts its policy to maximize the long-term reward. The long-term reward is either the sum of all future rewards or the discounted sum of future rewards to reduce the impact of future actions \cite{wiering2012reinforcement}.

RL is a method to solve sequential decision-making problems modeled as Markov Decision Process (MDP). MDPs can be continuous time, infinite horizon, partially observable, or a combination of these properties. Infinite horizon MDPs do not have a goal state, and the system runs forever \cite{wiering2012reinforcement}. The agent learns to maximize its total reward without expecting a goal state. In continuous-time MDPs, unlike discrete time, the decisions are made at every point of time. The formulation of continuous time MDP is similar to discrete MDP. However, these problems are more challenging to solve in general \cite{guo2009continuous}. Partially Observable MDP (POMDP) is a kind of MDP where the environment is partially observable for the agent due to different reasons such as limited sensors or uncertainty in the environment, and that limits the available information \cite{spaan2012partially,wiering2012reinforcement}. These variants of MDPs could also be modeled and solved by RL, although they need different considerations to define the states, in which the agent has a \textit{belief} about the environment rather than complete observation. In this article, we focus on discrete-time fully observable MDPs.

To model and solve a problem in the RL framework, different components of RL, including MDP modeling, algorithms, and hyper-parameters %, such as the number of training steps, the structure of policy function, etc., 
should be determined before starting the learning procedure. In practice, deciding on these components is rarely straightforward. It often requires considerable experience, repeated experimentation, and careful tuning through trial and error. Even small changes in hyper-parameters or network architecture can significantly affect performance, stability, and convergence. %As a result, 
Consequently, developing an effective RL solution can be time-consuming and highly sensitive to design choices. This is where AutoRL (Automated Reinforcement Learning) becomes particularly valuable. Instead of relying on manual tuning and expert intuition, AutoRL seeks to automate the selection and optimization of key elements. %such as algorithms, network structures, reward formulations, and training hyper-parameters. 
By systematically searching the configuration space, AutoRL can identify combinations that might not be obvious through manual experimentation. The practical importance of AutoRL lies in its ability to reduce the development effort while improving robustness and reproducibility. Automated configuration helps mitigate the risk of unstable training caused by poorly chosen parameters and makes RL methods more adaptable to new tasks and environments. This is especially important in real-world applications, such as robotics, autonomous systems, and complex optimization problems, where training instability or inefficient exploration can be costly. Moreover, by lowering the dependency on deep RL expertise, AutoRL makes reinforcement learning more accessible and easier to deploy across different domains.

A straightforward approach to state design is to use the environment observations directly as state representations and the agent's outputs as actions. While this is sufficient for some tasks, many problems require more informative state representations through preprocessing or feature extraction. 
%As a straightforward way of designing states, the exact observation from the environment is used as a state representation, and the agent's decision is directly sent to the environment as an action. The observations can be considered individually or stacked in batches. Although these states and actions might help the agent to find an optimal or near-optimal policy for some tasks, they are not necessarily the best representations for the states and the actions. In some other tasks, further processing of the environment is required to find a suitable state representation. 
As a simple example, normalization is usually necessary for the inputs of Neural Networks (NNs), and using raw observation of the environment may produce undesirable results \cite{sola1997importance}. NNs are strong function approximation tools for complex RL problems. Deep Reinforcement Learning (DRL) that combines RL and NN has been a breakthrough in AI. 
Furthermore, the effectiveness of an RL policy depends heavily on the choice of learning algorithm and hyperparameters, which are typically selected through expert knowledge and extensive experimentation.
%Furthermore, the performance of an RL policy is highly dependent on the RL algorithms and their hyper-parameters. Selecting algorithms and tuning hyper-parameters are normally done by using expert knowledge. However, many iterations are still needed to find the best set of hyper-parameters. 
In recent years, RL and DRL continue to expand into application domains  such as chemical engineering \cite{alabi2023automated} and optimization \cite{wu2025deep}. 
For example, within the optimization community, which studies how algorithms are developed to solve combinatorial optimization problems, various (deep) RL methods have been applied to learn heuristics to solve problems in transportation, manufacturing, logistics, etc \cite{mazyavkina2021reinforcement,zhang2025railcar,reijnen2026job}. % which have shown better performance than traditional algorithms in uncertain, dynamic environment \cite{}.  
%Recently, DRL has been popular for learning heuristics and solving different combinatorial optimization problems \cite{mazyavkina2021reinforcement}, including different variants of TSP. 
%and various RL methods have been applied to solve problems that are not classical RL applications, such as combinatorial optimization problems. 
In these new RL applications, expert knowledge of RL might not be present.  %addition, 
%In such non-ML there are many optimization problems in different areas 
%with no available expert knowledge on RL, that could be solved by RL.
These challenges motivate the need for AutoRL to automate key design decisions and make RL more accessible to non-experts. 
%Hence, automating an RL procedure to facilitate using this approach for non-experts and other research directions is of great importance.

\begin{figure*}[!t]
\centering
\begin{tikzpicture}[
  card/.style={draw, rounded corners=3.5pt, fill=white, inner sep=0pt},
  cardtitle/.style={font=\fontsize{9}{10}\selectfont\bfseries, anchor=north},
  mini/.style={draw=black!45, rounded corners=2pt, fill=white, align=left, font=\fontsize{8}{9}\selectfont, inner sep=2.8pt},
  plain/.style={draw=black!60, rounded corners=3.5pt, fill=black!3, align=center, font=\fontsize{9}{10}\selectfont, inner sep=4pt},
  lab/.style={font=\fontsize{8}{9}\selectfont, black!75},
  farr/.style={-{Stealth[length=2.6mm]}, very thick, black!70},
  aarr/.style={-{Stealth[length=2.4mm]}, thick, dashed, black!60},
  co/.style={draw=black!30, rounded corners=3.5pt, fill=black!2, align=left, font=\fontsize{8}{9.2}\selectfont, inner sep=4pt}]
% ---------- main row ----------
\node[plain, minimum width=18mm, minimum height=17mm] (prob) at (0.95,0) {\textbf{Problem}\\[-1pt]{\fontsize{8}{8.6}\selectfont sequential}\\[-2pt]{\fontsize{8}{8.6}\selectfont decision task}};
% MDP card
\node[card, draw=cMDP!75, minimum width=37mm, minimum height=34mm] (mdp) at (4.15,0) {};
\node[cardtitle, text=cMDP!60!black] at ($(mdp.north)+(0,-0.09)$) {MDP Modeling {\mdseries(Sec.~\ref*{sec:mdp})}};
\node[mini, text width=31mm, minimum height=7.5mm] at ($(mdp.center)+(0,0.66)$) {\textbf{State:} features,\\ embeddings};
\node[mini, text width=31mm, minimum height=7.5mm] at ($(mdp.center)+(0,-0.14)$) {\textbf{Action:} distributions,\\ discretization};
\node[mini, text width=31mm, minimum height=7.5mm] at ($(mdp.center)+(0,-0.94)$) {\textbf{Reward:} curriculum,\\ shaping};
% AutoRL search card
\node[card, draw=black!60, fill=black!2, minimum width=46mm, minimum height=34mm] (auto) at (8.75,0) {};
\node[cardtitle] at ($(auto.north)+(0,-0.09)$) {AutoRL configuration search};
\node[mini, draw=cALG!70, fill=cALG!6, text width=40mm, minimum height=7.5mm] at ($(auto.center)+(0,0.66)$) {\textbf{\color{cALG!60!black}Algorithm Selection} {\color{black!60}(Sec.~\ref*{sec:alg})}\\ bandits, dynamic configuration};
\node[mini, draw=cHPO!70, fill=cHPO!6, text width=40mm, minimum height=7.5mm] at ($(auto.center)+(0,-0.14)$) {\textbf{\color{cHPO!60!black}Hyper-param.\ Optim.} {\color{black!60}(Sec.~\ref*{sec:hyp})}\\ BO, bandits, EA, meta-gradient};
\node[mini, text width=40mm, minimum height=7.5mm] at ($(auto.center)+(0,-0.94)$) {\textbf{Cross-cutting:} L2L (Sec.~\ref*{sec:meta}),\\ NAS (Sec.~\ref*{sec:nn}), LLM (Sec.~\ref*{sec:llm})};
% training + evaluation
\node[plain, minimum width=24mm, minimum height=17mm] (train) at (12.55,0) {\textbf{RL Training}\\[-1pt]{\fontsize{8}{8.6}\selectfont agent--environment}\\[-2pt]{\fontsize{8}{8.6}\selectfont interaction}};
\node[card, draw=black!60, minimum width=26mm, minimum height=27mm] (eval) at (15.75,0) {};
\node[cardtitle] at ($(eval.north)+(0,-0.09)$) {Evaluation};
\node[mini, text width=20mm, align=center] at ($(eval.center)+(0,0.42)$) {return};
\node[mini, text width=20mm, align=center] at ($(eval.center)+(0,-0.27)$) {stability};
\node[mini, text width=20mm, align=center] at ($(eval.center)+(0,-0.96)$) {sample efficiency};
% ---------- solid RL pipeline arrows ----------
\draw[farr] (prob) -- (mdp);
\draw[farr] (mdp) -- (auto);
\draw[farr] (auto) -- (train);
\draw[farr] (train) -- (eval);
% ---------- dashed AutoRL outer loop (orthogonal, single bends) ----------
\draw[aarr] (eval.north) -- ++(0,0.75) -| (auto.north);
\draw[aarr] ($(eval.north)+(0,0.75)$) -| (mdp.north);
\node[lab, above] at (10.0,2.18) {AutoRL outer loop: evaluation feedback updates each automated component};
% legend
\node[lab, anchor=east, align=right] at (17.3,-1.8) {\tikz{\draw[farr] (0,0)--(0.5,0);}\, RL pipeline\quad \tikz{\draw[aarr] (0,0)--(0.5,0);}\, AutoRL outer loop};
% ---------- method cards ----------
\node[co, text width=46mm, anchor=north west] (cm) at (0.0,-2.15) {{\fontsize{8.5}{9.3}\selectfont\bfseries\color{cMDP!60!black}MDP Modeling (Sec.~\ref*{sec:mdp})}\\[1pt]\textbf{States:} tile coding \cite{sutton1996generalization}, embeddings \cite{dai2016discriminative}, aggregation \cite{afshar2020state}\\ \textbf{Actions:} Beta policy \cite{chou2017improving}, discretiz.\ \cite{tang2020discretizing}\\ \textbf{Rewards:} curriculum \cite{florensa2017reverse}, shaping \cite{ng1999policy}, EA hybrids \cite{khadka2018evolution}};
\node[co, text width=42mm, anchor=north west] (ca) at (5.15,-2.15) {{\fontsize{8.5}{9.3}\selectfont\bfseries\color{cALG!60!black}Algorithm Selection (Sec.~\ref*{sec:alg})}\\[1pt]bandit selection \cite{laroche2018reinforcement}\\ dynamic configuration \cite{biedenkapp2020dynamic,benjamins2024instance}\\ auxiliary-task selection \cite{vincent2024adaptive}};
\node[co, text width=52mm, anchor=north west] (ch) at (9.85,-2.15) {{\fontsize{8.5}{9.3}\selectfont\bfseries\color{cHPO!60!black}Hyper-parameter Optimization (Sec.~\ref*{sec:hyp})}\\[1pt]Bayesian \cite{hutter2011sequential,barsce2017towards,chen2018bayesian}; multi-fidelity \cite{li2017hyperband,parker2020provably}\\ population-based \cite{jaderberg2017population,franke2020sample}; evolutionary \cite{sehgal2019deep}\\ meta-gradient \cite{zahavy2020self}; RL-based \cite{jomaa2019hyp}};
\draw[cMDP!80, line width=2.2pt] ([xshift=0.5pt]cm.north west) -- ([xshift=0.5pt]cm.south west);
\draw[cALG!80, line width=2.2pt] ([xshift=0.5pt]ca.north west) -- ([xshift=0.5pt]ca.south west);
\draw[cHPO!80, line width=2.2pt] ([xshift=0.5pt]ch.north west) -- ([xshift=0.5pt]ch.south west);
\end{tikzpicture}
\caption{The AutoRL framework: solid arrows form the RL pipeline; dashed arrows form the AutoRL outer loop that feeds evaluation results back into the automated components. Bottom cards list representative methods.}
\label{fig:overview}

\vspace{-0.6em}
\end{figure*}

AutoRL provides a framework for automatically configuring key components of an RL pipeline, including MDP modeling, algorithm selection, and hyperparameter optimization. 
%to automatically make appropriate decisions about the settings of an RL procedure before starting the learning. In other words, RL components, including state, action and reward, algorithm selection, and hyper-parameters optimization, are determined through AutoRL, and the best configuration for each component is provided for an RL procedure to solve a task. 
Figure \ref{fig:overview} shows the AutoRL framework: the solid arrows form the standard RL pipeline from MDP modeling through configuration search to training and evaluation, and the dashed arrows form the AutoRL outer loop, which feeds evaluation results back into each automated component. The bottom cards list the reviewed technique families with representative methods. %To model and solve a problem using RL, the steps of this pipeline are followed, starting from MDP modeling. AutoRL aims to automate different steps of this pipeline and reduce the necessity of expert knowledge. 
AutoRL extends the idea of Automated Machine Learning (AutoML) \cite{HE2021106622} to reinforcement learning.
%We use AutoRL to emphasize the resemblance with Automated Machine Learning (AutoML) as a framework for automating supervised and unsupervised learning procedures. 
%According to \cite{HE2021106622}, an AutoML pipeline consists of the components of ML frameworks.  
AutoML starts with data preprocessing, followed by feature engineering and model generation, and ends with evaluation, whose output is used to reconfigure the earlier steps over several iterations until the optimal settings are derived. 
While AutoML automates tasks such as data preprocessing, feature engineering, model selection, and hyperparameter optimization, its methods cannot be directly transferred to RL because of the sequential nature of decision making and the high computational cost of policy evaluation. Consequently, AutoRL needs dedicated approaches to automate RL-specific design choices, reducing the dependency on expert knowledge, and improving the applicability of DRL to real-world problems. 
%Although some components of AutoML and AutoRL are similar, AutoML methods cannot necessarily be used in AutoRL because the configuration of the problems and the complexity of the evaluation step are different. 
%AutoRL pipeline builds a pipeline for RL to define and select components of RL automatically. In recent years, by combining Deep Learning and RL, the need for automating the components of DRL has been increased because many modeling decisions are made manually, and even RL experts have to test several different configurations related to state, action, reward, algorithm, and hyper-parameters, to obtain the best definition. Furthermore, experiments show that normally, there is a gap between the impressive results achieved by DRL algorithms in controlled environments and their practical applicability in real-world scenarios \cite{henderson2018deep}. AutoRL can address this gap by encouraging researchers to focus on aspects like sample efficiency, stability, and transferability, fostering the development of DRL techniques that are more applicable and effective in practice.

In this paper, we review relevant work that can be included in an AutoRL framework and elaborate on research challenges and directions in this relatively new research area. For each of the components mentioned above, different approaches in the literature are presented that might help automate the corresponding component. For example, approaches that modify the initial observations of the environment to define a state representation are candidate methods to use in automating states. AutoRL pipeline consists of modeling a particular problem as a sequential decision-making problem and MDP, selecting an appropriate algorithm, and tuning hyper-parameters. These three steps are illustrated in Figure \ref{fig:overview} followed by evaluation. Evaluating an RL algorithm is normally performed by tracing reward alteration during training and comparing the final total reward with some baselines. It means accumulating a sequence of rewards in the memory to use in the evaluation phase.

\begin{figure*}[!t]
\centering
\resizebox{\textwidth}{!}{\input{figure_taxonomy}}
\caption{Evolution of the AutoRL literature: one colored branch per automation target (and section), methods placed chronologically. Shaded bands mark eras; bold-border chips carry headline results quoted in the text.}
\label{fig:autorl-roadmap}
\end{figure*}

To demonstrate the purpose of AutoRL, take a classical optimization problem, Traveling Salesman Problems (TSP), as an example. %
The first step in solving TSP with DRL is to model the problem as a sequential decision making and determine MDP components. Vanilla TSP is defined as finding a tour with minimum length in a graph where all the nodes are visited exactly once. In a constructive solution approach, an agent starts from the source and walks through nodes until building a tour. At each timestep, the graph, current node, visited nodes, and remaining nodes are the observations of an environment. There are several ways like graph neural networks \cite{scarselli2008graph} and structure2vec \cite{dai2016discriminative} to convert the observations to state representation, and AutoRL helps to find the best approach among the possible methods for this conversion. The second step is to define an RL algorithm for updating the policy network. In actor-critic methods, the policy network and value network are the two main networks. These networks are updated during training. Popular RL algorithms such as A2C \cite{mnih2016asynchronous}, PPO \cite{schulman2017proximal}, ACER \cite{wang2016sample} and DQN \cite{mnih2015human} are available to train the policy network, and it is not easy to determine the best one for a problem; AutoRL aims to search among these candidates automatically. We focus on model-free RL algorithms where the transition probabilities between states are unknown. The third step is to set the hyper-parameters such as network architecture, learning rate, discount factor, etc. AutoRL employs hyper-parameter optimization methods in its framework to derive the optimal hyper-parameters for an algorithm and a set of problem instances.

Exploring AutoRL and relevant work is presented in \cite{parker2022automated}, which  covers a wide range of AutoRL techniques, including algorithm selection, hyperparameter tuning, and architecture design. 
While this survey \cite{parker2022automated} and our paper share a common focus on AutoRL, they differ in their specific approaches and coverage. For instance, the survey (\cite{parker2022automated}) provides a broad taxonomy that structures AutoRL around key components of the RL pipeline, such as hyperparameter optimization (HPO), where it discusses RL-specific methods like population-based training as well as Bayesian optimization tailored for noisy RL environments. Additionally, it covers neural architecture search (NAS) adapted for RL policies and value functions. %, with examples including evolutionary NAS for continuous control tasks, and addresses automated algorithm configuration and selection across off-policy and on-policy methods. 
Furthermore, the survey examines reward function design through techniques like inverse RL and motivated rewards, while also exploring environment design via unsupervised curriculum generation. In general, it encompasses both model-free and model-based RL paradigms. 
It also discusses key challenges such as long training times, sample inefficiency, evaluation, reproducibility, and deployment across domains including robotics, games, and molecular design. Furthermore, the survey highlights open research directions, such as standardized benchmarks, scalable optimization, and improved reproducibility. 
%Furthermore, the survey examines reward function design through techniques like inverse RL and motivated rewards, while also exploring environment design via unsupervised curriculum generation. In general, it encompasses both model-free and model-based RL paradigms, offering detailed discussions of challenges such as long training times, sample inefficiency, and evaluation in diverse domains such as games, robotics, and molecular design. %Spanning over 50 pages, the survey includes extensive references and case studies. 
%Moreover, it highlights open problems, for example, creating standardized benchmarks for AutoRL, improving reproducibility, scaling to high-dimensional real-world applications, and integrating multi-fidelity optimization. 
In contrast, our paper takes a more focused approach and provides a condensed overview of the key concepts in AutoRL, emphasizing model-free RL methods while largely omitting in-depth treatment of model-based aspects or specialized areas like automated reward and environment design. Besides, we extend our overview to AutoML, learning-to-learn, and automated neural network design techniques that can potentially be adopted for RL and assist RL designers in automating different RL components, offering insights into cross-domain adaptations not as prominently featured in the previous survey. In addition, we explore the integration of large
language models into AutoRL pipelines. Therefore, this paper serves as an overview and enlightens future directions rather than a pure summary of existing AutoRL techniques. % While both papers contribute to the understanding of AutoRL, the extent of their coverage, the level of detail, and the specific aspects emphasized vary, making them potentially complementary resources for those exploring automated reinforcement learning.

To provide a concise roadmap of the AutoRL landscape and illustrate how the subsequent sections relate to its key components, Figure~\ref{fig:autorl-roadmap} summarizes the major directions covered in this overview, including MDP modeling automation, algorithm/learner design, hyperparameter optimization, evaluation, and extensions such as meta-learning, neural architecture search, and LLM-based AutoRL techniques. Compared to \cite{parker2022automated}, 
%which provides a broad survey and taxonomy of AutoRL methods across the full pipeline, 
our paper is positioned as a focused overview centered on model-free RL. We emphasize the core automation knobs that practitioners most frequently tune in model-free DRL, while summarizing practical challenges that affect reliable automation, such as evaluation sensitivity, seed variance, and the cost of repeated full training. Moreover, we further highlight how ideas from AutoML, learning-to-learn, and neural architecture design can be adopted as reusable tools within the AutoRL outer loop, 
%using this perspective 
to motivate future research directions rather than exhaustively surveying all subfields.

This paper is organized as follows. In Section \ref{sec:mdp}, the works on automating the MDP modeling, including the definition of states, actions, and rewards, are reviewed. In Section \ref{sec:alg}, the process of RL algorithm selection is reviewed. Since algorithm selection is normally intertwined with hyper-parameter optimization, most of the combined work in algorithm selection and hyper-parameter optimization together with different hyper-parameter optimization work are presented in Section \ref{sec:hyp}. Section \ref{sec:meta} presents recent work in meta-learning that can be leveraged in an RL framework. In Section \ref{sec:nn}, previous work in optimizing and learning neural network architecture is reviewed. Section \ref{sec:llm} explores the integration of large language models into AutoRL pipelines. Section \ref{sec:limit} discusses limitations and future work in AutoRL. Section \ref{sec:ethics} discusses ethical considerations and potential risks associated with automating RL. Finally, Section \ref{sec:conc} concludes the paper. Throughout Sections \ref{sec:mdp}--\ref{sec:llm}, Table~\ref{tab:overview} categorizes the reviewed approaches along their key ideas, advantages, limitations, and applicable RL settings, Figure~\ref{fig:autorl-roadmap} organizes them chronologically, and the experimental results reported by state-of-the-art methods are quoted and analyzed where the corresponding methods are discussed.

\begin{table*}[!t]
\caption{Comparison of the reviewed AutoRL approaches, grouped by the automated component: key idea, advantages, limitations, and applicable RL settings.}
\label{tab:overview}
\centering
\fontsize{6.5}{7.2}\selectfont
\renewcommand{\arraystretch}{0.9}
\setlength{\tabcolsep}{2.5pt}
\begin{tabularx}{\textwidth}{@{}>{\raggedright\arraybackslash}p{2.5cm} Y Y Y >{\raggedright\arraybackslash}p{2.7cm}@{}}
\toprule
\textbf{Technique family} & \textbf{Key idea} & \textbf{Advantages} & \textbf{Limitations} & \textbf{Applicable RL settings} \\
\midrule
\multicolumn{5}{@{}l}{\emph{States (Sec.~\ref{sec:mdp})}}\\
\midrule
Expert feature construction \cite{sutton2018reinforcement,sutton1996generalization,sherstov2005function,abdoos2014hierarchical} & Polynomial features, coarse coding, and tile coding map observations to richer or discretized features & Cheap, robust, well understood & Number, shape, and placement of features chosen by the designer & Tabular/linear methods on low-dimensional spaces \\
Learned embeddings \cite{dai2016discriminative,vinyals2015pointer,scarselli2008graph,khalil2017learning,bello2016neural,echchahed2025state} & Neural encoders learn representations of structured instances end-to-end & Removes manual feature engineering; strong on graph/sequence problems & Restricted to structured inputs; encoder becomes a new design choice & DRL for graph or sequence tasks (e.g., COPs) \\
Adaptive discretization and aggregation \cite{whiteson2010adaptive,lin2010evolutionary,baumann2011state,afshar2020state} & Splitting heuristics, GAs, unsupervised quantization, or RL select the abstraction granularity & Automates the discretization resolution & Task-specific heuristics; fitness needs full RL runs; may discard reward-relevant distinctions & Value-based RL on continuous low-dimensional tasks; COPs \\
\midrule
\multicolumn{5}{@{}l}{\emph{Actions (Sec.~\ref{sec:mdp})}}\\
\midrule
Representation and distribution design \cite{tavakoli2020learning,stable-baselines,chou2017improving,tessler2019distributional} & Learn action representations or choose/learn the policy distribution (Gaussian, Beta, DPO) & Exploits action structure; Beta removes boundary bias; DPO learns beyond fixed families & Rank/family remain design decisions; extra machinery & Multi-dimensional or continuous action spaces \\
Discretization \cite{kanervisto2020action,tang2020discretizing,fourati2024stochastic} & Per-dimension grids with softmax; sampled subsets for large action sets & Makes value-based and on-policy methods applicable; scalable via sampling & Resolution trades precision against action-set size & Bounded continuous or very large discrete action spaces \\
\midrule
\multicolumn{5}{@{}l}{\emph{Rewards (Sec.~\ref{sec:mdp})}}\\
\midrule
Curriculum learning \cite{florensa2017reverse,ivanovic2019barc} & Grow the start-state distribution backwards from the goal & Mitigates sparsity without changing reward semantics & Requires a known goal state & Sparse-reward goal-reaching tasks \\
Bootstrapping and EA--DRL hybrids \cite{smart2002effective,khadka2018evolution,bodnar2020proximal,eysenbach2019search} & Initialize from demonstrations or evolutionary populations; decompose into subgoals & Improves exploration and stability under sparse feedback & Needs prior policies; sub-optimal priors bias values & Sparse-reward continuous control, long horizons \\
Reward shaping \cite{ng1999policy,grzes2017reward,zou2019reward,chiang2019learning,faust2019evolving,afshar2021reward,su2015reward,judah2014imitation,park2024shaping,ibrahim2024reward} & Add or learn auxiliary rewards (potential-based, meta-learned, parametric) & Densifies feedback; potential-based variants preserve the optimal policy & Shaping parameters need tuning; misspecification alters the task & Sparse or delayed-reward tasks \\
\midrule
\multicolumn{5}{@{}l}{\emph{Algorithm selection and hyper-parameter optimization (Secs.~\ref{sec:alg}--\ref{sec:hyp})}}\\
\midrule
Bandit-based selection and dynamic configuration \cite{degroote2016reinforcement,laroche2018reinforcement,vincent2024adaptive,biedenkapp2020dynamic,benjamins2024instance} & Treat algorithms/configurations as bandit arms or a contextual MDP; adapt online & No full training per candidate; per-instance, time-varying configurations & Portfolio fixed a priori; generalization depends on training instances & Episodic RL; iterative algorithms \\
Bayesian optimization and multi-fidelity bandits \cite{hutter2011sequential,barsce2017towards,chen2018bayesian,beeks2022deep,jamieson2016non,li2017hyperband,parker2020provably,yuan2025ultho,becktepe2024arlbench} & Surrogates propose configurations; successive halving allocates budget adaptively & Sample-efficient per trial; order-of-magnitude cheaper than full evaluations & Full RL run per evaluation; low fidelities may mis-rank & Any RL task; large search spaces \\
Gradient- and RL-based tuning \cite{maclaurin2015gradient,zahavy2020self,jomaa2019hyp,dong2018hyperparameter,vuong2019pick,white2016greedy,chen2020automatic,souza2024autorl} & Meta-gradients through training, a learned configuration policy, or an amortized mapping & Tunes within a single run; dynamic and amortized configurations & Only differentiable hyper-parameters; meta-training cost; noise & Actor-critic DRL; amortized HPO across tasks \\
Evolutionary and population-based \cite{fernandez2018parameters,sehgal2019deep,franke2020sample,jaderberg2017population,stanley2019designing} & Populations evolve hyper-parameters (and networks) during training, sharing experience & Online schedule adaptation; strong reported gains without algorithmic changes & Population-scale compute; sensitive to noisy fitness & Off-policy DRL with shared replay \\
\midrule
\multicolumn{5}{@{}l}{\emph{Learning to learn (Sec.~\ref{sec:meta})}}\\
\midrule
Learned optimizers, learned RL algorithms, in-context RL \cite{andrychowicz2016learning,wang2016learning,daniel2016learning,xu2017reinforcement,moeini2025survey,polubarov2025vintix} & Replace the update rule or whole RL algorithm with a learned sequence model & Adapts to task distributions; no fine-tuning at test time (in-context) & Bound to the training distribution; large offline cost & Meta-RL over related tasks \\
Meta-learning and discovered update rules \cite{finn2017model,zou2019reward,xu2018meta,xu2020meta,kirsch2019improving,oh2020discovering} & Learn initializations, objectives, returns, or full update rules from experience & Can surpass hand-designed components & Noisy meta-gradients; overfitting to training environments & DRL with differentiable meta-objectives \\
Environment design, resets, transfer \cite{dennis2020emergent,eysenbach2018leave,souza2024transfer} & Generate curricula, learn reset policies, transfer configurations across tasks & Reduces manual environment/curriculum design & Adds a second learning problem & Sim-based training; task sequences \\
\midrule
\multicolumn{5}{@{}l}{\emph{Neural network architecture (Sec.~\ref{sec:nn})}}\\
\midrule
RL-based NAS \cite{zoph2016neural,zoph2018learning,baker2016designing,miao2022differentiable,cassimon2024scalable,rahmani2025nasdqn} & A controller generates architectures rewarded by validation performance & Removes manual architecture design; transferable cells reduce cost & Compute-hungry; proxy evaluations may mis-rank & Policy/value network design \\
Neuroevolution and efficiency search \cite{miikkulainen2019evolving,stanley2019designing,elthakeb2019releq,garcia2019a} & Evolve topologies and hyper-parameters; RL selects per-layer bit-widths & Jointly optimizes structure and weights; hardware-aware & Population compute; task-specific reward design & Gradient-free settings; resource-constrained deployment \\
\midrule
\multicolumn{5}{@{}l}{\emph{LLM for AutoRL (Sec.~\ref{sec:llm})}}\\
\midrule
Reward design \cite{yu2023l2r,qu2025latentreward,ma2024eureka,chen2025elemental,lu2025rewarddiscovery} & Translate intent into reward code, refined via evaluate--refine loops & Removes the main manual bottleneck; can surpass expert rewards & Reward-hacking risk; ambiguous text-only specifications & Continuous control and robotics \\
Algorithm evolution and recipe proposal \cite{oh2025discorl,yang2024opro,zhang2023llmhpo,eimer2023hprl} & Discover update rules; LLM proposals warm-start outer-loop search & Extends automation beyond tuning to algorithm design & Prompt sensitivity; proposals need ground-truth validation & Limited-budget HPO; algorithm design \\
MDP automation \cite{wang2024lesr} & Generate state code, validators, and environment wrappers & Injects domain priors; improves sample efficiency and portability & Generated code may be inconsistent or unsafe & Domains with structured context \\
LLM as policy learner \cite{li2022plmidm,zhang2023rememberer,zhao2024expel,wang2024voyager,yan2025llmprior,du2023ellm} & LLM backbone with memory, skills, or action priors, improved by RL-style feedback & Continual improvement without fine-tuning; better exploration & High inference cost per decision; grounding failures & Language-rich interactive tasks \\
\bottomrule
\end{tabularx}
\end{table*}

\section{Markov Decision Process Components}
\label{sec:mdp}

Formally speaking, a discrete-time finite horizon MDP is a tuple $(S,A,r,T,\gamma)$, where $S$ is the state space, $A$ is the action space, $r:S\times A\rightarrow r \in \mathbb{R}$ is an immediate reward value denoting the benefit of transition from current state $s_t\in S$ to the next state $s_{t+1}\in S$, and $\gamma$ is the discount factor \cite{puterman2014markov}. At each decision moment or discrete timestep $t$, an agent interacts with the environment, and its goal is to learn a policy $\pi : S \times A \rightarrow \pi \in [0,1]$ that determines a probability value for each action. Following a greedy, $\epsilon$-greedy, softmax, or other action selection policies, the agent takes an action according to the probabilities and transitions to the next state. In other words, the agent observes a state $s_t\in S$ and performs an action $a_t\sim\pi(\cdot\mid s_t)$. Taking an action has two consequences. First, the agent receives a reward value $r_t$. Then, the state of the environment transitions from $s_t$ to a new state $s_{t+1}$ based on the transition probability $T$. The agent updates the policy during the learning to find the optimal policy that yields the maximum total reward. RL provides an interaction-based framework to solve the MDP and learn the policy $\pi$. The AutoRL needs to define the following four components in RL: state, action, reward, and transition probability. The transition probability is mostly unknown for model-free RL problems \cite{sutton2018reinforcement}. Hence, we focus only on state, action, and reward definitions in this section.

\subsection{Methods for Automating States}

In many classical RL task, such as mountain car, cart pole, and pendulum \cite{wiering2012reinforcement}, the agent’s raw observation is treated as the state. However, this observation is not always an efficient representation. In applications with continuous or high-dimensional spaces, the state space can be extremely large, making value function approximation challenging. As a result, learning effective mappings from observations to compact state representations has received significant attention. Existing approaches can be grouped into two categories: (1) methods that rely on expert-designed transformations of raw observations, where hyperparameter tuning and configuration remain manual, and (2) methods that automatically learn state representations, reducing dependence on expert knowledge.

\subsubsection{Transferring raw observations to state representation}
\label{sec:state-rep}

Manipulating raw observation and constructing new features are widely used for deriving state representation. These methods range from simple approaches such as tile coding applied to linear function approximation methods for problems like n-state random walk \cite{sherstov2005function,abdoos2014hierarchical}, to more complex methods like structure2vec \cite{dai2016discriminative} and Pointer Networks \cite{vinyals2015pointer} for graph Combinatorial Optimization Problems (COPs) such as Vertex Cover \cite{khalil2017learning} and TSP \cite{bello2016neural}. They are mainly employed for expanding the observation to more useful representations with or without taking the final policy into account. For problems like combinatorial optimization \cite{mazyavkina2020reinforcement}, robot navigation \cite{altuntacs2016reinforcement}, and real-world business problems like train shunting \cite{peer2018shunting} and online advertising \cite{afshar2019reinforcement,Afshar2023}, raw observation of the environment may require processes to derive state representation.

Each state is represented by an n-dimensional observation vector, whose elements are scalar features. For complex observations, such as images, the data can be flattened into feature vectors. However, in many problems, the raw observations are insufficient to represent the state effectively. 
%Each state consists of a n-dimensional observations vector from the environment, and each observation is a scalar value called a feature. In other words, the observations in an n-dimensional observation space are feature vectors with n entries. In more complex tasks like image classification, where the observation is a matrix rather than a vector, they can be flattened to vectors. Exact observations from the environment in some problems are not sufficient for representing states. 
For instance, when two observation dimensions interact (e.g., their signs jointly determine which actions are appropriate), a raw two-entry vector does not encode this interaction. Since the observations in many environments are represented by numerical values, features can be interpolated to generate new meaningful features. One of the simplest families of features used for interpolation is \textit{Polynomial Features} \cite{sutton2018reinforcement}. Polynomial features are obtained by modeling the observations as any order-n polynomial. Formally speaking, assume $O=(o_1,o_2,...,o_k)$ is an observation vector from the environment. The new state $s_i$ corresponding to an observation $o_i$ is defined as: $s_i=\prod^{k}_{j=1}{o_j^{c_{i,j}}}$,  
where $c_{i,j}$ is an integer denoting the degree of $j^{th}$ term of observation in the definition of the $i^{th}$ term of state representation. 
% Although higher-order polynomials entail more complex functions, the number of features grows exponentially when $k$ increases. Therefore, a subset of features is selected in most cases. 
This approach is mainly used for deriving state representation of linear function approximation algorithms when the important interactions between the features are not included in the observations of the environment.

\textit{Coarse Coding} is another useful approach for generating features, especially when the observation of the environment is not informative enough \cite{sutton1996generalization}. For example, assume a task with two-dimensional state space where each region of the space has its own characteristics. In order to capture pertinent information about the environment, coarse coding introduces some overlapping circles whose status shows the corresponding state of observation. Each observation lies in one or more circles, and circles are called present/absent or active/inactive based on the location of the observation. If the observation lies in $i^{th}$ circle, its corresponding value in state $s$ is 1; otherwise, it is 0. Using this method, the feature vector of states extends to $n$ binary values. For example, with $n=6$ overlapping circles, an observation that lies in the third and the sixth circles is encoded as the feature vector $(0,0,1,0,0,1)$.

\textit{Tile Coding} \cite{sutton2018reinforcement} is a widely used approach for converting continuous space to discrete which %makes its purpose similar to Coarse coding.
is easier to manage and reduces the complexity of the problem.
In tile coding, $n$ \textit{tiling} that each has a fixed number of \textit{tiles} are offset from each other by a uniform amount in each direction, and the tiles containing the observation across all tilings constitute its binary features.

With the popularity of Deep Neural Networks (DNNs) in the past few years, feature engineering is mainly performed by DNNs, and researchers focus more on designing the DNNs. Nevertheless, for some problems like COPs, processing the raw representation of problem instances in order to derive effective state representation improves the quality of the solution. 
% For this reason, much research focuses on developing methods for transforming the original observation of the environment to state representation. 
In \cite{dai2016discriminative}, an approach named structure2vec for representing structured data like trees and graphs is introduced. This approach is based on the idea of embedding latent variable models into feature space. A vector for representing a graph is obtained by employing probabilistic kernels to find latent variable models, and a neural network is trained to output the embedding of a graph based on nodes' attributes. This idea is used in \cite{khalil2017learning} for solving graph-based combinatorial optimization problems such as Minimum Vertex Cover and TSP. The graph embedding network is learned by fitted Q-learning, and the output of the network is used as a greedy policy to incrementally create the solution to the problem.

\subsubsection{Automatically defining the state representation}

Some approaches in the literature try to automatically find state representation. Adaptive tile coding \cite{whiteson2010adaptive} begins with a single tile covering the entire state space and uses two heuristics to guide splitting. The first monitors the minimum Bellman error and triggers a split when improvement stalls. The second examines action-selection conflicts within a tile and splits it if conflicts exceed a threshold. To allow uneven splits, \cite{lin2010evolutionary} employs a genetic algorithm to automatically learn tile codings and state abstractions for large state spaces. Starting from a single tile, the GA determines when and where to split. Each individual represents a tiling encoded as a binary decision tree, and its fitness is measured by the performance of an RL agent using that tiling. Mutation operators either shift existing split boundaries or introduce new splits. This method is demonstrated on the Mountain Car and Pole Balancing continuous control tasks \cite{lin2010evolutionary}.

A common strategy for large state spaces is state aggregation, which groups states with similar characteristics to reduce complexity. In \cite{baumann2011state}, state aggregation for Q-learning in continuous domains is achieved by combining Growing Neural Gas (GNG) with Q-learning. GNG incrementally builds a topological representation from codeword vectors that quantize the state space, refining it during training and thereby learning state abstractions automatically. Similarly, \cite{afshar2020state} extends tile coding to discretize continuous state spaces for solving the knapsack problem. A single 
$n$-dimensional tiling converts continuous item values into discrete representations, where each item corresponds to a dimension. Reinforcement learning is then used to automatically determine the number of tiles per dimension. In this formulation, states correspond to items, and actions determine the number of tiles. A recent survey \cite{echchahed2025state} organizes modern representation-learning objectives for deep RL, including metric-based, self-predictive, and reconstruction-based methods, into a unified taxonomy, indicating that learned-representation methods have largely displaced hand-designed feature construction in high-dimensional tasks.

% Focusing on graph-based state representation leads to other promising works. In \cite{perozzi2014deepwalk}, a representation named \textit{DeepWalk} is introduced that generates a sequence of nodes using random walk. This model is used in natural language processing and trains a model similar to word2vec for representing a graph. In \cite{zhou2017scalable}, an asymmetric proximity preserving graph embedding based on random walk is developed that can be used to represent a graph for social network problems like link prediction.

\subsubsection{Challenges}

Approaches for transforming observations to state representation have parameters and settings that properly tuning them would significantly increase the total reward. For instance, although tile coding and coarse coding are useful approaches for handling large and continuous state space, the number of tilings, tiles, and circles have to be determined by the system designer or expert. Furthermore, promising NN-based methods like Pointer Network require appropriate information about the problem instance, which is still the task of the RL designer. Therefore, completely replacing expert knowledge with automation levels is a challenging task in defining state representation. The other challenge is the generalization of proposed methods. Methods such as structure2vec apply only to graph-based problems, and adaptive tile coding likewise requires task-specific adaptations. Deriving a generic state representation method would improve an AutoRL pipeline drastically, which is not well studied in the literature.

\subsection{Methods for Automating Actions}
\label{sec:auto-action}

In many RL tasks, actions are mainly the decisions of the agent that alter the state of the environment. Different types of actions, such as continuous, discrete, multi-dimensional, bounded, or unbounded, would entail policies with different qualities. For example, a continuous action like prices in a dynamic pricing task can be modeled as either continuous or discrete action space. On one hand, continuous actions might be more precise; however, it is not possible to model them with tabular reinforcement learning or function approximation approaches like DQN that consider an output for each action. On the other hand, although discrete actions are easier to model, modeling a continuous space as discrete might be tricky, especially when small changes in the action would have a large impact on the total reward. Therefore, deriving a proper action representation is very important as it is difficult to find the best representation for actions that end up with the best policy. For this reason, automating the definition of action spaces is necessary for many tasks. Action spaces could be a combination of discrete and continuous for multi-dimensional spaces like robot joints in robot navigation problems. For many continuous control tasks such as a pendulum or BipedalWalker \cite{brockman2016openai}, the continuous action space could be discretized to represent discrete actions. In this subsection, we review learning actions and discretizing continuous action space separately.

\subsubsection{Learning actions}

Action representation learning in order to improve action values and policy has become popular in recent years. In \cite{tavakoli2020learning}, the action representation of multi-dimensional action spaces is learned using hyper-graph. Hyper-graph is the generalization of a graph in which each single hyper-edge could contain one or more vertices. In this modeling, the actions are modeled as vertices in hyper-graph, and the goal is to learn the representation of hyper-edges. Each hyper-edge has a parametric function that maps the state to values over its action combinations, and a fixed mixing function aggregates these vectors into Q values; the hyper-graph itself is selected through a rank $r$ that bounds the cardinality of its hyper-edges.

An efficient way to represent actions is to model the output of the policy network as a continuous Probability Density Function (PDF). In common practice, Gaussian distribution is used for the policy, and mean and standard deviation are learned during the training \cite{stable-baselines}. Gaussian distribution has been successful in many tasks with continuous action space. However, infinite support of this distribution might introduce bias in policies obtained from policy gradient algorithms. To solve this issue, Beta distribution is used as the policy PDF instead of Gaussian distribution in \cite{chou2017improving}. The authors show that using Beta PDF for policy reduces bias whilst the performance is not negatively affected.

One approach to automatically derive the policy distribution is introduced in \cite{tessler2019distributional},  %According to this work, 
which states that the policy gradient update rule with parametric distribution functions results in sub-optimal policies. This sub-optimality is in the distribution space, and learning the policy distribution is a solution. For this purpose, distributional policy optimization (DPO) is presented as an update rule that minimizes the distance between the policy and a target distribution, as %shown below: %:. This update rule is shown in Equation \ref{eq:DPO-GAC}.
%\begin{equation}
$$\pi_{k+1}=\Gamma(\pi_k-\alpha_k \nabla_\pi d(\mathcal{D^{\pi_{k}}_{\mathcal{I}^{\pi_{k}}}},\pi)|_\pi=\pi_k),$$ 
%\label{eq:DPO-GAC}
%\end{equation}
where $\Gamma$ is a projection operator onto the set of distributions, $d$ is a distance measure, $\mathcal{D}^{\pi}_{\mathcal{I}^\pi}$ is a distribution over all states, and actions that their advantage value is positive. In order to minimize the distance between two distributions, Implicit Quantile Network \cite{dabney2018implicit} is employed by using the Wasserstein distance metric. 

\subsubsection{Discretizing continuous actions}

Continuous action spaces are challenging in many tasks. As mentioned before, some RL algorithms like DQN do not work well for continuous action spaces because they rely on $\epsilon$-greedy algorithm, and the best action is required at each step. Finding the best action in a continuous space needs an optimization step for each interaction with the environment, which is intractable. For discrete action spaces, a separate output is considered in the policy network for each action, which is not possible when the action space is continuous, as it needs an infinite number of outputs. Although discretization \cite{kanervisto2020action} makes the action space discrete and manageable, it is not suitable for tasks that are very sensitive to small alterations of actions.

Sometimes, continuous action spaces could be transformed to discrete while retaining necessary information for action selection as shown in \cite{tang2020discretizing} for on-policy control RL when the domain of all the continuous actions is between -1 and 1. The set of discrete actions for each dimension is $\mathcal{A}_i=\{\frac{2j}{K-1}-1\}_{j=0}^{K-1}$, where $K$ is the number of discrete actions. The discrete policy is a neural network that outputs a logit $L_{ij}$ for $j^{th}$ action in $i^{th}$ dimension. For each dimension $i$, the logits are combined through a softmax function to compute the probability of choosing each action. This approach is integrated with TRPO and PPO to be evaluated on MuJoCo benchmarks \cite{tang2020discretizing}. A complementary difficulty is that discretization can produce very large discrete action sets. Stochastic Q-learning \cite{fourati2024stochastic} addresses this by evaluating only a random sub-linear subset of actions at each step, keeping value-based methods tractable when the number of discrete actions grows.

\subsubsection{Challenges}

Similar to state representation, approaches for determining action space contain hyper-parameters, and finding appropriate configurations is important. One way to optimize the parameters is through the hyper-parameters optimization module, as depicted by a dashed arrow between actions and hyper-parameter optimization in Figure \ref{fig:overview}. This is challenging because there might be several action spaces, and each has some parameters. Hence, the search space is relatively large and the optimization procedure is computationally expensive. On the other hand, finding the optimal action space needs many trial and error steps by checking different definitions. The automation level may help decide the policy distribution or discretization approach for continuous actions, which is necessary. Designing the structure of the policy output is normally performed using expert knowledge, which is not always available. This is an interesting research direction that may influence positively on the output of an RL framework.

\subsection{Automated Reward Function}
\label{sec:auto-reward}

The reward function is a key component of an MDP that strongly affects policy quality. For example, in a 2-D grid world, different reward designs yield different behaviors. If only reaching the goal provides reward, the agent may wander since intermediate moves are not penalized. In contrast, a reward based on distance to the goal encourages efficient paths. Designing an effective reward function is therefore crucial and typically requires expert knowledge, often involving trial and error. Automating reward design can help agents discover better policies more efficiently. In general, three main approaches to reward design can be considered for automation, allowing the agent to search for the most suitable representation.

\subsubsection{Curriculum Learning}

Curriculum learning is useful for training in environments with sparse rewards. In many tasks, such as robot navigation, the search space is large and only the goal state yields a positive reward. Curriculum learning addresses this by starting training from states near the goal and gradually increasing difficulty as the initial states are placed farther away. In \cite{florensa2017reverse}, a dynamic programming-based curriculum learning method is proposed to address reward sparsity. The approach begins training from states near the goal and continues until the agent demonstrates mastery. Then, new start states are generated via random walks from previously learned states, gradually expanding the search space. The so-called \textit{reverse learning} strategy effectively mitigates sparse reward challenges.
Similarly to \cite{florensa2017reverse}, the backward learning method in \cite{ivanovic2019barc} starts from the state in the vicinity of the goal state and increases the distance when the agent demonstrates mastery in solving the problem. Unlike \cite{florensa2017reverse}, where the state space expands by random walk, the new states are obtained by computing an approximate backward reachable set, which represents all points in the state space that the agent is able to reach a certain region in a fixed and short amount of time. 
% The main contribution of this work is to update the state space in a way that only reachable states from the goal state are preserved.

Although curriculum learning is helpful in solving sparsity, its main application is in goal-searching tasks. Using curriculum learning in COPs is quite challenging. Learning heuristics in these problems might not be adapted with curriculum learning because the optimal solutions are not usually available in advance. Hence, curriculum learning in a RL pipeline is limited to goal-based tasks where the goal state is available.

\subsubsection{Bootstrapping}

Bootstrapping methods start learning from a pre-defined policy. This policy could be either for a similar task or designed by a human. 
% Bootstrapping serves as a way of developing the reward function in the literature. 
A typical approach based on bootstrapping is introduced in \cite{smart2002effective}, where the learning process is split into two phases. In the first phase, the robot is controlled by an existing control policy or directly by a human. In the latter case, the robot is navigated by humans in the environment, and it updates the value function during this phase without changing the policy. The second phase is a typical reinforcement learning process, and the robot updates the policy based on the value function, which is initialized in the first phase. 
% Bootstrapping has some similarities with transfer learning. They both involve leveraging existing knowledge to improve performance in a new task. They both aim to enhance model performance by utilizing pre-existing information or parameters, either through resampling techniques in bootstrapping or by transferring knowledge from a source domain to a target domain in transfer learning.
In \cite{khadka2018evolution}, a hybrid EA-DRL algorithm is proposed to address reward sparsity, poor exploration, and unstable convergence. Each individual in the evolutionary population is a DNN actor within the Deep Deterministic Policy Gradient (DDPG) framework \cite{lillicrap2016continuous}. A population of actors is initialized, and each actor’s fitness is defined as the cumulative reward. New generations are produced via crossover, which randomly exchanges weight segments between parents, and mutation, which adds Gaussian noise to network weights. Meanwhile, experiences collected by all actors are stored in a replay buffer and used to train a separate DDPG actor. 
% This trained actor is periodically injected back into the evolutionary population. Unlike standard DDPG, the main actor leverages experiences from multiple policies, making the approach a form of bootstrapping that exploits shared information to improve learning. 
\cite{bodnar2020proximal} extends the evolutionary RL framework of \cite{khadka2018evolution} by introducing personal replay buffers and redesigned crossover and mutation operators. Each individual maintains a small genetic memory that stores its recent experiences. In the proposed crossover, called Q-filtered distillation, the child’s genetic memory is populated with recent experiences from its parents. The child is initialized with one parent’s weights and trained on this memory, optimizing a loss that blends the parents’ policies. The new mutation operator, proximal mutation, refines Gaussian perturbation by scaling the noise with the summed gradients over a batch of transitions, reducing destructive updates and improving stability.
% This approach is evaluated on MuJoCo environments, including Hopper, Ant, and HalfCheetah.
% In sparse-reward tasks with high-dimensional state spaces, learning long-horizon policies is challenging. Planning methods are effective when distances between states and local policies are easy to define, but they struggle with high-dimensional observations. Conversely, RL handles high-dimensional inputs well but often fails on long-horizon tasks. 
% To combine the strengths of both, 
\cite{eysenbach2019search} aims to tackle long-horizon RL tasks. A task is decomposed into simpler goal-reaching subtasks, where identifying subgoals is formulated as a shortest-path problem using a learned distance metric. By assigning a reward of $-1$ per step until the goal is reached, the Q-function aligns with shortest-path objectives. 
% A directed graph is constructed over states stored in the replay buffer, with nodes representing observations and edges weighted by their distances. The algorithm then computes the shortest path from the start to the goal and guides a goal-conditioned policy through successive intermediate subgoals.

\subsubsection{Reward Shaping}

Reward shaping involves designing a proxy reward function to maximize expected return. While rewards are often straightforward in tasks like video games, they require careful design in domains such as robot navigation \cite{kim2023automated} and multi-objective combinatorial optimization problems (COPs) \cite{ilavarasi2014variants}. 
% For instance, in goal-search tasks, a reward must balance objectives like obstacle avoidance, goal reaching, and maintaining a minimum distance. 
% Similarly, in a TSP with time windows and prizes, the reward should encourage minimal tour length, maximal prize collection, and adherence to time constraints. 
In such cases, poorly chosen rewards can drastically affect performance, making careful reward design essential. The following sections review examples and applications of reward shaping across different domains.

% In \cite{judah2014imitation}, reward shaping in  imitation learning is explored. 
In imitation learning, the agent cannot obtain any reward. State-action pairs from a target policy are demonstrated to the agent, and the goal is to mimic the target policy using these state-action pairs. In \cite{judah2014imitation}, a shaped reward function is provided to the agent, which is not necessarily aligned with the target policy. The policy is a parametric function with parameters updated by maximizing the reward. In \cite{su2015reward}, reward shaping is studied for Spoken Dialogue Systems (SDS) modeled as a POMDP. Because SDS typically provide only a final reward at the end of a dialogue, the reward signal is sparse. To address this, domain knowledge is used to generate an additional reward via a recurrent neural network (RNN) trained with supervised learning on annotated data. 

% The RNN takes belief state–action pairs as input and predicts the dialogue return. The agent then combines this auxiliary reward with the environment feedback to learn its policy.

% In \cite{su2015reward}, the benefit of reward shaping in Spoken Dialogue Systems (SDS) as a Partially Observable MDP (POMDP) is studied. In SDS, an agent interacts and converses with the clients, and a numerical reward value is received at the end of the conversation. Since there is no intermediate reward, the reward function is sparse. In this work, domain knowledge is used to provide another reward signal. This extra reward value is obtained from a recurrent neural network (RNN), which is trained by supervised learning using previously annotated data. The belief of actual state in the POMDP context and action pairs are the inputs of the RNN, and the scalar return of the dialogue is the target value. The agent uses the extra reward value along with the feedback of the environment to learn its policy.

A well-known reward shaping approach is potential-based reward shaping \cite{ng1999policy}. Let $r(s,a,s')$ be the immediate reward of taking action $a$ in state $s$ and going to state $s'$. Potential-based reward shaping employs a function $F(s,a,s')=\gamma\Phi(s')-\Phi(s)$ where, $\gamma$ is discount factor, and $\Phi(s)$ and $\Phi(s')$ are potential functions of states $s$ and $s'$, respectively. The new reward function is $r(s,a,s')+F(s,a,s')$, which is the sum of the original reward and the potential-based reward shaping. In \cite{grzes2017reward}, potential-based reward shaping is analyzed in episodic RL. 
% The authors show that temporal credit assignment or reward value could speed up training if a potential-based reward function is added to the original reward.
% An important requirement of this work is that the policy derived by shaped reward should be equivalent to the original policy. 
The authors found that potential-based reward shaping preserves the optimal policy when the goal states are predefined terminal states and the shaped reward is zero in the goal states. They observe that policy invariance is violated for finite horizon domains with multiple terminal states, and propose to set the potential value of terminal states as zero to solve this issue. Learning a potential function $\Phi(s)$ using meta-learning is studied in \cite{zou2019reward}. The potential function is defined by a neural network, and its parameters are updated using the Model-Agnostic Meta-Learning (MAML) algorithm \cite{finn2017model}. 
% The update equations of MAML are as follows.
% \begin{equation}
% \label{eq:objective-p2p1}
% \phi_i=\theta-\alpha\nabla_\theta\mathcal{L}_{\tau_i}(f_\theta)
% \end{equation}
% \begin{equation}
% \label{eq:objective-p2p2}
% \theta=\theta-\beta\nabla_\theta\mathbb{E}_{\tau_{i}}\mathcal{L}_{\tau_i}(f_{\phi_i})
% \end{equation}
% where $\tau_i$ is a MDP task deriving from a particular task distribution, $\theta$ is the parameters of the value network, $\phi_i$ is the parameters of potential-based reward shaping which is task-specific, $\alpha$ and $\beta$ are learning rates, $f(\theta)$ and $f(\phi_i)$ are the value network and reward shaping network respectively, and $\mathcal{L}$ is the loss function. 
The parameters of the networks are trained through running an adapted version of DQN with replay memory.
In \cite{chiang2019learning}, learned proxy reward functions are proposed for path-following and target-based robot navigation to address sparse binary rewards (i.e., goal is reached or not). The method alternates between two stages: (1) optimizing the parameters of the reward function, and (2) training fixed-architecture actor-critic networks with the learned reward. The reward parameters that yield the highest objective value are selected, and RL is used to train the final actor-critic policy. In \cite{faust2019evolving}, the authors learn a parametric function for typical continuous control RL problems such as \textit{Ant}, \textit{Walker2D}, and \textit{Humanoid} \cite{openai-gym}. For each problem, a particular parametric reward function is defined, and the same algorithm is used to learn both the parameters of the reward and the policy network. Actor-critic algorithms, including Proximal Policy Optimization (PPO) \cite{schulman2017proximal} and Soft Actor-Critic (SAC) \cite{haarnoja2018soft}, are used with parametric reward function, and the method outperforms the same algorithms without parametric reward on the aforementioned tasks. In \cite{afshar2021reward}, a reward vector is introduced to optimize reserve prices in real-time ad auctions with sparse binary feedback. The agent uses a policy network to map ad slot features to a distribution over reserve prices, while the environment only indicates whether the slot is sold. To provide a richer signal, the reserve price range is divided into weighted sub-intervals, and the reward is calculated as the inner product of the weight vector with an interval-based reward vector that activates the entry for the chosen interval.

\subsubsection{Challenges}
As mentioned earlier, curriculum learning is well suited for goal-searching problems but is less applicable to path-following problems, where the goal state is unknown. In optimization problems, leveraging solutions from simpler instances to solve more complex ones remains a challenging but promising research direction for AutoRL. 
%As mentioned before, Curriculum learning is appropriate for goal-searching problems. This method fails for path-following problems where the goal state is unknown. Using this method in AutoRL is challenging because the goal state is mainly the solution to the optimization problems. However, the solution to easy instances of a problem might help to solve complex instances, and this generalization is relatively hard but is important progress in AutoRL. 
Bootstrapping methods require initial policies that are provided either by expert knowledge or by following other learning or optimization methods. Though expert knowledge is helpful, it is not available for most problems. Besides, using any policy other than optimal policy does not help the learning because that biases the value function, and a sub-optimal policy is learned. Nevertheless, finding the optimal approach is time-consuming, and a level of automation could be largely beneficial. Reward shaping methods have some parameters that are tuned before starting the training phase. Similar to parametric state and action representation methods, these hyper-parameter settings highly influence the total reward. Tuning the hyper-parameters is challenging and time-consuming, which makes it an interesting research direction. Indeed, recent position work argues that automating environment and reward shaping, rather than improving policy-optimization algorithms, is now the main bottleneck for practical RL, especially in sim-to-real robotics \cite{park2024shaping}, and a recent overview catalogs reward-engineering practice across application domains \cite{ibrahim2024reward}.
Another interesting research direction is about the effects of intrinsic rewards and whether they help in solving the problem of sparse rewards \cite{singh2004intrinsically}. For example, giving a reward when the agent has a lot of options available or when the agent encounters a new situation.

\section{Automated Algorithm Selection}
\label{sec:alg}

When the problem is modeled as a sequential decision-making problem, and the components of MDP are defined, the next step in solving it with RL is to select an appropriate algorithm. One way to reduce the search space is to filter the algorithms based on the class of the problem. For example, if the states and actions are discrete and finite, tabular RL algorithms like typical Q-Learning and SARSA \cite{sutton2018reinforcement} are suitable candidates, and there is no need to search over the class of sophisticated algorithms in DRL. Moreover, if the model of the environment is known, a wide variety of model-based algorithms like dynamic programming could be utilized. Despite having different contexts, algorithm selection approaches in AutoML can provide insights for AutoRL \cite{raschka2018model}. Most of the work in the domain of RL algorithm selection is intertwined with hyper-parameter optimization. For this reason, we explain combined algorithm selection and hyper-parameter optimization methods in the next section and present a few works only on algorithm selection in this section.

In \cite{degroote2016reinforcement}, algorithm selection in supervised learning is modeled as a contextual multi-armed bandit problem. Each decision step observes the dataset's meta-feature vector as the bandit context, and UCB or $\epsilon$-greedy learns the value of each candidate arm across datasets. The algorithm selection problem in episodic RL tasks is modeled as a multi-armed bandit problem in \cite{laroche2018reinforcement} to decide which RL algorithm is in control for each episode. At each step, an algorithm from a given portfolio is selected to control the episode, following Epochal Stochastic Bandit Algorithm Selection: the time scale is divided into epochs of exponential length, and the policies of the algorithms are updated only at epoch boundaries, which handles the non-stationarity induced by learning. In \cite{vincent2024adaptive} Adaptive-Network (A-Net), an approach to enhance deep reinforcement learning by enabling agents to select targets dynamically is introduced. A-Net learns an auxiliary task selection policy that adapts based on the current training phase, improving the sample efficiency and overall performance of reinforcement learning models. It addresses the challenge of target selection in multi-task environments, where choosing the right auxiliary task is critical for learning efficiency. The authors demonstrate that A-Net outperforms baseline methods across multiple environments, showing more effective and efficient learning behaviors.

\textbf{Challenges.} Generally speaking, the RL algorithm can be classified according to the type of policy or value function. Through this categorization, the policy or value function could be either tabular or parametric. One initial challenge in RL algorithm selection is to decide between these two. If the state and action spaces are relatively small, tabular methods are more appropriate, whereas these methods do not work for large and continuous space and action spaces. Discretization is another solution that is discussed in section \ref{sec:auto-action}.
After identifying an appropriate class of RL algorithm, selecting an algorithm to learn the policy is another challenge. One way is to treat the algorithm as a hyper-parameter and optimize that in the hyper-parameters optimization module, which is normally followed in AutoML. However, this approach requires a particular hyper-parameters optimization framework because the quality of the algorithm depends on the problem, its MDP modeling, and parameter settings. Furthermore, the number of required timesteps varies for different algorithms, and this makes comparing the algorithms challenging. In sum, selecting the proper RL algorithm for a task is difficult, and it highly depends on the problem.

\section{Hyper-Parameter Optimization}
\label{sec:hyp}

An optimal RL configuration for solving a sequential decision-making problem highly depends on promising hyper-parameters settings \cite{bischl2023hyperparameter}. Hyper-parameters are fixed during the training, and they are usually set by RL experts prior to starting the interactions, although they might vary at different points of time \cite{mohan2023autorl}. For example, the learning rate in policy gradient or value function update formula, discount factor, eligibility trace coefficient, and parameters of a parametric reward shaping method are hyper-parameters. Different tasks require different sets of hyper-parameters, which makes hyper-parameter optimization challenging, and automating this process would be very useful. Many different approaches are developed for automatically optimizing hyper-parameters of supervised learning algorithms that could be adapted with RL to optimize hyper-parameters of RL algorithms. In this section, we first review hyper-parameters optimization approaches and then their applications. At last, the main challenges of optimizing hyper-parameters are elaborated.

\subsection{Methods for tuning hyper-parameters}

This subsection presents the previous hyper-parameters optimization work categorized by their core methodology. Figure~\ref{fig:hpomech} illustrates the mechanisms of the three search paradigms that dominate hyper-parameter optimization for RL: sequential model-based optimization, multi-fidelity budget allocation, and online population-based training.

\begin{figure*}[!t]
\centering
\begin{tikzpicture}[
  card/.style={draw=cHPO!75, dashdotted, thick, rounded corners=2pt, fill=white},
  ptitle/.style={anchor=west, font=\fontsize{8.5}{9.5}\selectfont\bfseries},
  pm/.style={font=\fontsize{7.5}{8.5}\selectfont\itshape, black!70, anchor=west},
  pl/.style={font=\fontsize{7.5}{8.8}\selectfont, black!85, anchor=west},
  bx/.style={draw=black!70, rounded corners=2pt, align=center, font=\fontsize{7.5}{8.4}\selectfont, inner sep=2.6pt, fill=white},
  hbx/.style={draw=cHPO!80, rounded corners=2pt, align=center, font=\fontsize{7.5}{8.4}\selectfont, inner sep=2.6pt, fill=cHPO!8},
  sb/.style={draw=cHPO!80!black, circle, fill=white, inner sep=0.8pt, font=\fontsize{6.5}{7}\selectfont\bfseries, text=cHPO!70!black},
  pa/.style={-{Stealth[length=1.8mm]}, thick, black!70}]
% =============== (a) SMBO loop (official cycle) ===============
\begin{scope}[shift={(0,0)}]
\draw[card] (-0.35,-1.62) rectangle (5.25,3.3);
\node[ptitle] at (-0.15,3.0) {(a) Sequential model-based optimization};
\node[bx, text width=21mm] (init) at (1.3,2.58) {initial design};
\node[bx, text width=21mm, minimum height=8mm] (arch) at (1.3,1.7) {archive\\[-2pt]$\{(\theta_j, f(\theta_j))\}$};
\node[hbx, text width=21mm, minimum height=8mm] (fit) at (3.9,1.7) {fit surrogate\\[-2pt]model $M$};
\node[hbx, text width=21mm, minimum height=8mm] (acq) at (3.9,0.45) {maximize\\[-2pt]$\alpha_M(\theta) \rightarrow \theta_{i+1}$};
\node[bx, text width=21mm, minimum height=8mm] (ev) at (1.3,0.45) {evaluate $f(\theta_{i+1})$\\[-2pt](one full RL run)};
\draw[pa] (init) -- (arch);
\draw[pa] (arch) -- (fit);
\draw[pa] (fit) -- (acq);
\draw[pa] (acq) -- (ev);
\draw[pa] (ev) -- (arch);
\node[sb] at (2.6,1.94) {1};
\node[sb] at (4.14,1.07) {2};
\node[sb] at (2.6,0.2) {3};
\node[pl] at (-0.15,-0.62) {\textcircled{\tiny 1} fit surrogate $M$ on the archive};
\node[pl] at (-0.15,-0.98) {\textcircled{\tiny 2} propose $\arg\max_\theta \alpha_M$\ \ \textcircled{\tiny 3} evaluate, repeat};
\node[pm] at (-0.15,-1.36) {SMAC \cite{hutter2011sequential}, RLOpt \cite{barsce2017towards}, BO for AlphaGo \cite{chen2018bayesian}};
\end{scope}
% =============== (b) successive halving rungs (official blocks) ===============
\begin{scope}[shift={(6.05,0)}]
\draw[card] (-0.35,-1.62) rectangle (5.25,3.3);
\node[ptitle] at (-0.15,3.0) {(b) Multi-fidelity successive halving};
% rung 1: 8 configs
\foreach \k in {0,...,3}{\fill[cHPO!45, rounded corners=1pt] ({0.35+\k*0.44},0.25) rectangle ({0.67+\k*0.44},0.57);}
\foreach \k in {4,...,7}{\fill[black!22, rounded corners=1pt] ({0.35+\k*0.44},0.25) rectangle ({0.67+\k*0.44},0.57);
  \node[font=\fontsize{7}{7}\selectfont, black!60] at ({0.51+\k*0.44},0.41) {$\times$};}
\node[pl, font=\fontsize{7}{7.8}\selectfont, black!75] at (3.95,0.41) {$8$ @ $b$};
% rung 2: 4 configs
\foreach \k in {0,...,1}{\fill[cHPO!60, rounded corners=1pt] ({0.35+\k*0.44},0.95) rectangle ({0.67+\k*0.44},1.27);}
\foreach \k in {2,...,3}{\fill[black!22, rounded corners=1pt] ({0.35+\k*0.44},0.95) rectangle ({0.67+\k*0.44},1.27);
  \node[font=\fontsize{7}{7}\selectfont, black!60] at ({0.51+\k*0.44},1.11) {$\times$};}
\node[pl, font=\fontsize{7}{7.8}\selectfont, black!75] at (2.2,1.11) {$4$ @ $2b$};
% rung 3: 2 configs
\fill[cHPO!80, rounded corners=1pt] (0.35,1.65) rectangle (0.67,1.97);
\fill[black!22, rounded corners=1pt] (0.79,1.65) rectangle (1.11,1.97);
\node[font=\fontsize{7}{7}\selectfont, black!60] at (0.95,1.81) {$\times$};
\node[pl, font=\fontsize{7}{7.8}\selectfont, black!75] at (1.35,1.81) {$2$ @ $4b$};
% rung 4: winner
\fill[cHPO, rounded corners=1pt] (0.35,2.35) rectangle (0.67,2.67);
\node[star, star points=5, fill=white, inner sep=0.7pt] at (0.51,2.51) {};
\node[pl, font=\fontsize{7}{7.8}\selectfont, black!75] at (0.9,2.51) {$1$ @ $8b$};
% promotion arrows
\draw[pa] (1.15,0.62) -- node[right, font=\fontsize{7}{7.6}\selectfont, cHPO!75!black]{keep top $1/2$, double budget} (1.15,0.9);
\draw[pa] (0.72,1.32) -- (0.72,1.6);
\draw[pa] (0.51,2.02) -- (0.51,2.3);
\node[sb] at (5.0,0.41) {1};
\node[sb] at (4.72,0.76) {2};
\node[sb] at (2.08,2.51) {3};
\node[pl] at (-0.15,-0.62) {\textcircled{\tiny 1} sample $n$ configs with budget $b$ each};
\node[pl] at (-0.15,-0.98) {\textcircled{\tiny 2} rank, keep top $1/2$, double $b$\ \ \textcircled{\tiny 3} repeat to one};
\node[pm] at (-0.15,-1.36) {Hyperband \cite{li2017hyperband}, PB2 \cite{parker2020provably}, ULTHO \cite{yuan2025ultho}};
\end{scope}
% =============== (c) PBT lineage diagram (official DeepMind schematic) ===============
\begin{scope}[shift={(12.1,0)}]
\draw[card] (-0.35,-1.62) rectangle (5.25,3.3);
\node[ptitle] at (-0.15,3.0) {(c) Online population-based training};
% worker lineages over training time
\draw[black!55, very thick] (0.6,2.1) -- (4.65,2.1);
\draw[black!55, very thick] (0.6,1.45) -- (4.65,1.45);
\draw[black!55, very thick] (0.6,0.8) -- (2.0,0.8);
\node[font=\fontsize{7.5}{8}\selectfont, black!60] at (2.17,0.8) {$\times$};
% exploit fork from the best lineage + explore perturbation
\draw[cHPO, very thick] (2.0,2.1) -- (2.55,0.95) -- (4.65,0.95);
\node[star, star points=5, fill=cHPO, draw=white, line width=0.4pt, inner sep=1.5pt] at (2.55,0.95) {};
% start / end markers
\foreach \y in {2.1,1.45,0.8}{\draw[black!60, fill=white, thick] (0.6,\y) circle (1.5pt);}
\foreach \y in {2.1,1.45}{\fill[black!60] (4.65,\y) circle (1.6pt);}
\fill[cHPO] (4.65,0.95) circle (1.6pt);
% annotations in clear zones
\node[font=\fontsize{7}{7.6}\selectfont, cHPO!80!black, anchor=east, align=right] at (5.05,2.52) {exploit:\\[-1pt]fork from the best};
\draw[black!35, densely dotted] (3.45,2.42) -- (2.15,2.05);
\node[font=\fontsize{7}{7.6}\selectfont, cHPO!80!black, anchor=west] at (2.95,0.5) {explore: perturb $\theta$};
\draw[black!35, densely dotted] (2.92,0.58) -- (2.62,0.88);
\node[font=\fontsize{7}{7.6}\selectfont, black!60, anchor=west] at (0.6,0.42) {pruned worker};
\draw[black!35, densely dotted] (1.55,0.52) -- (2.08,0.72);
\draw[-{Stealth[length=1.7mm]}, thin, black!65] (0.6,-0.1) -- (4.65,-0.1);
\node[font=\fontsize{7}{7.6}\selectfont, black!70, below] at (2.6,-0.14) {training time (single run)};
\node[sb] at (0.28,2.1) {1};
\node[sb] at (2.06,1.66) {2};
\node[sb] at (2.85,1.2) {3};
\node[pl] at (-0.15,-0.62) {\textcircled{\tiny 1} train a population of workers in parallel};
\node[pl] at (-0.15,-0.98) {\textcircled{\tiny 2} exploit: fork from best\ \textcircled{\tiny 3} explore: perturb $\theta$};
\node[pm] at (-0.15,-1.36) {PBT \cite{jaderberg2017population}, SEARL \cite{franke2020sample}, STAC \cite{zahavy2020self}};
\end{scope}
\end{tikzpicture}
\caption{The three dominant HPO search paradigms in RL, in the algorithm forms of their original sources: (a) the SMBO loop, (b) successive-halving promotion rungs, and (c) PBT population snapshots with exploit and explore. Circled numbers mark the steps listed in each panel.}
\label{fig:hpomech}
\end{figure*}

\subsubsection{Gradient Descent}
Backpropagation is a main method for training neural networks in which the gradient of the loss function is computed with respect to the weights. This gradient is propagated backward through the network, and the new weights are obtained by a variant of the gradient descent algorithm. The hyper-parameters of gradient descent with momentum, including decay rate and learning rate, are included in the backpropagation algorithm, and they are optimized together with neural network weights in \cite{maclaurin2015gradient}. In \cite{zahavy2020self}, meta-gradient descent is used to automatically adapt hyperparameters online. The algorithm is leveraged to self-tune parameters of the actor-critic loss function. On Atari-57, the resulting self-tuning agent (STAC) raises the median human-normalized score of an IMPALA baseline from $243\%$ to $364\%$ at 200M frames \cite{zahavy2020self}.

Regression models are used in \cite{souza2024autorl} to develop AutoRL-Sim, a simulation environment designed to address combinatorial optimization tasks such as the Traveling Salesman Problem (TSP), Asymmetric TSP (ATSP), and Sequential Ordering Problem (SOP). AutoRL-Sim automates reinforcement learning (RL) processes using AutoML techniques to optimize parameters like learning rate and discount factor, improving solution accuracy and efficiency. The simulator is built in R, is freely available, and supports post-experiment analysis with graphical outputs, offering users flexibility through both predefined and customizable modules.

\subsubsection{Bayesian Optimization}
Bayesian Optimization methods, including Sequential Model-based Algorithm Configuration (SMAC), have been very popular in AutoML \cite{hutter2011sequential}. These methods are beneficial for optimizing expensive to-evaluate functions such as the performance of supervised learning algorithms. The idea of Bayesian Optimization is extended to RL hyper-parameters optimization in \cite{barsce2017towards}. In this work, the \textit{RLOpt} framework integrates Bayesian optimization with RL: a Gaussian-process surrogate maps hyper-parameters to observed performance and proposes the next setting from the history of (hyper-parameter, performance) tuples (Figure~\ref{fig:hpomech}a). In \cite{chen2018bayesian}, tuning hyper-parameters of Alpha-Go method \cite{silver2016mastering} using Bayesian Optimization shows improvement in playing strength. Specifically, Bayesian optimization improved AlphaGo's self-play win rate from $50\%$ to $66.5\%$ during its development \cite{chen2018bayesian}. In \cite{beeks2022deep}, the authors develop a DRL approach to solve a multi-objective order batching problem, where the weights of two objectives in the reward function are tuned with Bayesian Optimization.

\subsubsection{Multi-Armed Bandit}
A bandit-based hyper-parameter optimization algorithm named \textit{Hyperband} is proposed in \cite{li2017hyperband}. This algorithm is based on \textit{SuccessiveHalving} \cite{jamieson2016non}, which repeatedly evaluates the remaining configurations and discards the worst half until a single configuration remains (Figure~\ref{fig:hpomech}b). On supervised-learning workloads, Hyperband provides over an order-of-magnitude speedup compared to random search and Bayesian-optimization competitors \cite{li2017hyperband}, although transferring such multi-fidelity gains to RL is not automatic because early-training performance in RL can mis-rank final performance.
In \cite{parker2020provably}, a method is introduced for hyper-parameter optimization using population-based bandits, ensuring provable efficiency in an online setting. By combining bandit algorithms with a population-based approach, the method aims to dynamically adapt to changing optimization landscapes, demonstrating improved efficiency in hyper-parameter tuning.

\subsubsection{Evolutionary Algorithms}
In \cite{fernandez2018parameters}, the parameters of $SARSA(\lambda)$ and $Q(\lambda)$ as two RL algorithms based on eligibility traces are optimized using Genetic Algorithm (GA). In this method, a vector containing all hyper-parameters is considered as a \textit{chromosome}, and the mutation and crossover are performed on this vector. The algorithm is tested on an under-actuated pendulum swing-up, and the authors show that the selected parameters maximize the end performance. Another application of GA for optimizing the hyper-parameters of RL algorithms is presented in \cite{sehgal2019deep}, where the parameters of Deep Deterministic Policy Gradient (DDPG) with Hindsight Experience Replay (HER) \cite{andrychowicz2017hindsight} are learned through GA. Chromosomes concatenate binary encodings of the discount factor, polyak-averaging coefficient, actor and critic learning rates, exploration rate, and noise parameters, and fitness is the inverse of the number of epochs needed to approach the maximum success rate.

Hyper-parameter optimization for DRL faces three difficulties: optimal settings change over the course of training, testing each candidate requires a full training run, and dynamic modification of the network is rarely considered. A joint optimization approach based on evolutionary algorithms that optimizes the agent's network and its hyper-parameters simultaneously is presented in \cite{franke2020sample}. In the evolutionary framework, each individual is a DRL agent consisting of a policy and a value network together with RL algorithm parameters. Rollouts of each individual are stored in a shared replay memory to be used as experiences for other agents. Each agent is evaluated by running for at least one episode in the environment, and the mean reward of the agent is used to measure its fitness. After crossover and mutation, all the agents in the environment are trained using the experiences in the shared replay memory. This approach is applied to the TD3 algorithm in the MuJoCo continuous control benchmark. Compared with population-based training (PBT) \cite{jaderberg2017population}, which tunes hyper-parameters online with a population of concurrent learners and, e.g., raised the human-normalized score of an unchanged UNREAL agent on DeepMind Lab from $93\%$ to $106\%$, this approach (SEARL) reduces the environment interactions required for meta-optimization by up to an order of magnitude \cite{franke2020sample}. This integration of evolutionary algorithms and neural networks is known as neuroevolution \cite{stanley2019designing}.

\subsubsection{Greedy Algorithms}
To optimize the decay rate in  eligibility trace algorithms such as $TD(\lambda)$, \cite{white2016greedy} proposes a greedy algorithm that defines $\lambda$ as a function of states for RL algorithms with linear function approximation. 
During policy evaluation, $\lambda$ is selected greedily based on the weight vector, current and next state observations, immediate reward, and importance sampling to minimize the error between the resulting return and the Monte Carlo return ($\lambda=1$).

\subsubsection{Reinforcement Learning}
Hyper-parameters optimization is modeled as a sequential decision-making problem in \cite{jomaa2019hyp}, and RL is used to find the optimal hyper-parameters. In this framework, the agent learns to explore the space of hyper-parameters of a supervised learning algorithm, and the final parameters minimize error on the validation set. 
This method,  
originally designed for supervised learning, % and works based on training and validation datasets. However, the general idea 
can be extended to RL algorithms. 
States are the dataset meta-features plus the history of evaluated hyper-parameters and their performance, actions assign hyper-parameter values, and the reward is the resulting performance; a DQN \cite{mnih2015human} then learns a policy that determines optimal hyper-parameters for each dataset.

Normally, the hyper-parameters of an algorithm are optimized once, and they are fixed during the entire run of the algorithm. However, because most of AI algorithms are iterative, the optimal hyper-parameters might change over time. The problem of dynamic algorithm configuration is studied in \cite{biedenkapp2020dynamic}, and RL is used to derive a policy for optimal configuration in each step. In this modeling, states are descriptions of an algorithm A, and actions are assigning particular values to hyper-parameters of A. The reward function depends on the instances drawn from the same contextual MDP. The optimal policy is obtained either by a tabular Q-learning or DQN to select the configuration that has the highest discounted reward. Since instances' information is part of the state, the optimal configuration might be different for different instances. Follow-up work shows that the generalization of such learned configuration policies depends strongly on which training instances they are exposed to, and proposes automated instance selection to improve generalization \cite{benjamins2024instance}.

\subsubsection{Neural Networks}
One main challenge of using well-known hyper-parameter optimization in AutoML, such as sequential model-based optimization (SMBO) and SMAC, is the time needed to perform necessary iterations. These iterative algorithms take a remarkable amount of time to optimize the hyper-parameters, and this is highly prohibitive in the RL framework. This challenge is the motivation of developing a neural network for finding a mapping between data and hyper-parameters \cite{chen2020automatic}. The meta-features of the dataset are the input of a Convolutional Neural Network (CNN), and the hyper-parameters of the algorithm are the output. Training the CNN is based on supervised learning using subsets of a large dataset as the training data. The target hyper-parameters used for supervised learning are obtained by Bayesian Optimization.

% \subsection{Applications}

In \cite{dong2018hyperparameter}, hyper-parameters optimization for object tracking algorithms is modeled as RL and Normalized Advantage Functions \cite{gu2016continuous} is used to learn a policy network that receives a state and returns the optimal hyper-parameters settings for a particular object tracking algorithm. The state combines the tracker's search-region heat map, its parameters, and appearance features, and the reward is the tracking accuracy.

One important research direction in robotics is domain randomization. Domain randomization trains over a distribution of simulated MDPs to obtain a policy for the real environment; because fixed distribution parameters are not always adequate, cross-entropy search is used to learn them in \cite{vuong2019pick}. In this method, the policy parameters function is a function of MDP distribution parameters, and optimal policy parameters are derived by PPO. The MDP parameters are acquired by maximizing the discounted return of following the optimal policy in a real environment.

Application of RL in different domains requires special consideration for tuning the hyper-parameters. For instance, in \cite{ottoni2020tuning}, RL is leveraged to solve the Sequential Ordering Problem (SOP) - a variant of TSP with a precedence constraint. Tuning parameters in this work is performed by testing different RL algorithms, including SARSA and Q-learning, different reward definitions, and some different values for $\epsilon$ in $\epsilon$-greedy. This configuration of the algorithm is used to solve SOP.

\textbf{Challenges.}
Although several methods are developed to tune hyper-parameters, hyper-parameter optimization for RL algorithms may be computationally expensive because the agent needs to interact with an environment and update its policy continuously for a number of timesteps or until reaching convergence. This is usually a time-consuming and intractable process that needs special considerations. Efficient methods in AutoML, such as Bayesian Optimization, may work well for RL models with small states, actions, and trajectories. However, for complex tasks with large state and action space or long trajectories, existing methods require considerable adaptation to provide the best configuration in a reasonable time. This is a significant challenge in RL hyper-parameter optimization.

One main difference between supervised learning and reinforcement learning is the evaluation criteria because there is no target value like supervised learning datasets to show the desired behavior. Instead, an agent seeks to find a compromise between exploration and exploitation that helps as hints \cite{barsce2017towards}.

Multi-fidelity methods \cite{forrester2007multi} do not run the ML for the full budget for every hyper-parameter but only for a limited budget (low fidelity). Then, only promising hyper-parameters are run for longer (high fidelity). Adapting this method to AutoRL is challenging and also interesting because it might reduce the required time budget for optimization.

Recent work has started to address both the cost and the comparability problems of hyper-parameter optimization in RL. ARLBench \cite{becktepe2024arlbench} provides a standardized and computationally efficient benchmark for evaluating HPO methods on RL algorithms such as PPO, DQN, and SAC, while ULTHO \cite{yuan2025ultho} performs ultra-lightweight online hyper-parameter selection within a single training run using a clustered multi-armed-bandit formulation.

\section{Learning to Learn}
\label{sec:meta}

Apart from the three main components of an RL framework, levels of automation are also presented in the literature for the procedures that can be placed in more than one component. For example, the gradient descent method is used for updating the parameters of parametric functions like policy or reward. In this section, recent works on automating these kinds of procedures are reviewed. Most of these works are inherently developed for supervised learning. However, the same motivation and requirements hold for RL, which shows interesting research directions for future work.

In \cite{andrychowicz2016learning}, the normal gradient descent formula is replaced with a new formula in which a function of gradient value is used rather than the original gradient in the update rule: % The new gradient descent update equation is %shown in Equation (\ref{eq:new-gradient}).
\begin{equation}
\label{eq:new-gradient}
\theta_{t+1}=\theta_t+g_t(\nabla{f}(\theta_t),\phi)
\end{equation}
where $\theta$ is the parameter of the objective function $f(\theta)$. In this formulation, $g_t(\nabla{f}(\theta_t),\phi)$ is a function of the gradient of $f$ with parameters $\phi$ which is obtained by a recurrent neural network. The method consists of two neural networks. Function $f$ known as \textit{optimizee} is represented by a feed-forward neural network with parameters $\theta$. The gradients of $f$ are used in function $g$, which is represented by an LSTM recurrent neural network. These gradients plus the hidden states of RNN are the input, and $g$ is the output, which is used in the update rule shown in Equation (\ref{eq:new-gradient}). The method is tested on a class of 10-dimensional quadratic functions and also on MNIST and CIFAR-10 datasets, and the results show the power of using RNN for $g$.

Selecting RL algorithms is normally performed with expert knowledge or using approaches explained in previous sections. Instead of using existing RL algorithms that perhaps each works well on a certain type of problem, a model is introduced in \cite{wang2016learning} that can learn an RL algorithm. Specifically, a distribution $D$ over MDPs is defined, and an RL algorithm is learned in the sense that it performs well on the MDPs drawn from $D$. During solving MDP with reinforcement learning, an RNN is trained that its inputs are states, actions, and rewards of the MDP, and the output is the policy. Therefore, a recurrent neural network works as a RL algorithm. This idea has recently re-emerged at scale as \emph{in-context RL}, where transformer-based agents distill the policy-improvement process itself and adapt to unseen tasks purely by conditioning on interaction histories, without any parameter update \cite{moeini2025survey}. For example, Vintix \cite{polubarov2025vintix} learns a single cross-domain action model in this way.

Neural Networks are mainly trained using a variant of Stochastic Gradient Descent (SGD) such as normal SGD, SGD with momentum, or Adam. The performance of these algorithms depends on the selected learning rate, which would be different for different contexts and applications. An automatic framework based on RL for deriving the best learning rate is proposed in \cite{daniel2016learning}. In this approach, a set of features is introduced to represent the states in the RL modeling. These features include the variance and the gradient of the loss function. The state representation is used to train a policy using the Relative Entropy Policy Search (REPS) algorithm \cite{peters2010relative} for deciding the learning rate of a particular optimizer. REPS ensures the policy updates are close to each other by constraining the updates through a bound on Kullback-Leibler (KL) divergence.

A general meta-learning approach that can be applied to any model learned by gradient descent, is presented in \cite{finn2017model}. The goal of this approach is to update the model's parameters using a few training steps to produce acceptable results on a new task. For this purpose, a parametric model is defined that aims to work well on tasks drawn from a task distribution. The algorithm starts with random initialization of model parameters. During each iteration, a set of tasks is sampled from the given distribution, and the tasks' adapted parameters are updated using gradient descent on a particular number of examples. At the end of each iteration, the model's parameters are updated using the adapted parameters. The paper discusses the application of this method in supervised learning and classification, and a possible extension to RL is also explained. 
In \cite{oh2020discovering}, a meta-learning approach is presented to discover an entire update rule for reinforcement learning algorithms based on a set of environments.

In \cite{eysenbach2018leave}, a reset policy is considered together with the reinforcement learning policy to reset the environment prior to an expensive-to-reset situation. A reset policy is necessary for tasks where manual resets are costly, such as an autonomous car crashing at high speed. In \cite{eysenbach2018leave}, the off-policy actor-critic method is used to learn policies, where the $Q$ values of the main policy and the reset policy are jointly learned. The reset policy takes over selecting an action to abort the episode if its Q value for a particular action taken by the forward policy is lower than a threshold. The safe actions are the reversible sequence of actions where the agent can always undo them.

Meta-gradient descent has recently been explored in defining components of reinforcement learning. In \cite{xu2020meta}, meta-gradient descent is used to discover a customized objective function parametrized by a neural network, while \cite{xu2018meta} applies gradient-based meta-learning to optimize return estimation and its decay rates.  In \cite{kirsch2019improving}, a meta-learning approach leverages the experiences of many complex agents to learn a low-complexity neural objective function that decides the learning method of future individuals. RL environments play a crucial role in training and evaluation of RL agents. %provide a framework for training, testing, and evaluating RL agents. Different algorithms work differently in various environments, and modeling an environment and its settings 
%are of high importance in RL design. 
In \cite{dennis2020emergent}, unsupervised environment design is used to automatically generate diverse tasks that promote skill acquisition and knowledge transfer. 
%the generation of complex tasks through unsupervised environment design is explored. The authors propose a method for creating environments that foster the emergence of diverse skills without explicit supervision, facilitating effective knowledge transfer across tasks.
Furthermore, 
\cite{souza2024transfer} proposes Auto\_TL\_RL, which integrates transfer learning (TL) with autoRL, to enable knowledge transfer across combinatorial optimization problems such as the Asymmetric Traveling Salesman Problem (ATSP) and the Sequential Ordering Problem (SOP) while automatically selecting RL hyperparameters, such as the learning rate and discount factor.  This significantly reduces computational time and improves solution quality. %In this paper AutoRL (Automated Reinforcement Learning) is utilized to automatically select the best configurations for reinforcement learning parameters, such as the learning rate and discount factor. AutoRL simplifies the process of conducting RL experiments by reducing manual intervention in parameter tuning.

\textbf{Challenges.} Learning-to-learn for RL is still challenging in practice because it must generalize across tasks while remaining stable and affordable to train. Meta-objectives can easily overfit to a narrow task distribution and transfer poorly when tasks, dynamics, or reward scales shift. Meta-training is also computationally expensive, since each outer-loop update requires many inner-loop rollouts and policy updates, and the cost grows quickly with long horizons and high-dimensional observations. Moreover, the inherent stochasticity of RL can make meta-gradients noisy and unstable, so multi-seed evaluation and consistent experimental protocols are often necessary to avoid selecting brittle meta-initializations.

\section{Automating Neural Network Architecture}
\label{sec:nn}

Combining DNNs and RL introduces several successful algorithms for solving complex problems like COPs and video games. Although using DNNs improves the quality of function approximation, the performance of the DRL algorithms highly depends on the proper structure of DNNs. Different methods have been proposed in the literature for automatically defining the best DNN structures. These methods can be categorized as hyper-parameters optimization; however, we assign a separate section to emphasize their importance.

In \cite{zoph2016neural}, a recurrent neural network - the controller - is trained with reinforcement learning where the outputs of this RNN determine the architecture of another neural network - the child network - that is used for prediction. The child network is trained on a dataset, and its performance is the reward for training the controller; although developed for supervised learning, the approach extends to other domains.

According to \cite{zoph2018learning}, applying the method presented in \cite{zoph2016neural} directly on large datasets is computationally expensive. The solution introduced in \cite{zoph2018learning} is to search on a proxy dataset, which is rather small, and then transfer the learned network architecture to a large dataset. The search space contains generic convolutional cells expressed as repeated motifs, and the final architecture is a combination of two cell types whose feature maps have the same and half dimensions, respectively.

Deriving DNN architecture can be modeled as a sequential decision-making problem, and RL is a suitable approach for solving that. In \cite{baker2016designing}, a meta-modeling algorithm based on RL named MetaQNN is introduced to generate CNN architecture. The process of CNN architecture selection is automated by a Q-learning agent whose goal is to find the best CNN architecture for a particular machine-learning task. The validation accuracy of the given ML dataset is used as the reward value for the agent, and the actions are obtained by following the $\epsilon$-greedy algorithm and exploring in a discrete and finite space of layer parameters. This approach shows high performance for image classification tasks. RL is also used as a neural architecture search technique for RL in \cite{miao2022differentiable}. More recent work both scales RL-based NAS controllers to transferable, single-shot search \cite{cassimon2024scalable} and applies architecture search directly to the value-network design of DRL agents themselves \cite{rahmani2025nasdqn}.

Optimizing the parameters of NNs is an interesting research area as those parameters greatly influence the performance. In \cite{xu2017reinforcement}, a DRL approach is presented for automatically learning the value of the learning rate in stochastic gradient descent. Given the model parameters of the neural network and the training samples, the authors use the actor-critic policy gradient method to pick a learning rate through the policy network for the gradient descent algorithm. The state in this modeling is a compact vector of the model parameters to avoid processing all the parameters of large networks. Immediate reward for updating the learning rate generator network is the difference between the loss function of the main model in two consecutive time steps. According to the presented results, automatically deriving the learning rate increases the quality of the prediction model.

In \cite{garcia2019a}, the authors focused on the problem of lifelong learning and how an agent learns the optimal policy of a particular MDP using the information of a sequence of MDPs from the same distribution. Specifically, the problem in this work is to search for an optimal exploration policy that an agent follows during exploration in the environment. Each agent maintains two policies: an exploitation policy, which is task-specific, and an exploration policy, which is shared between all the MDPs drawn from the same distribution. At each time step, each of the two policies provides an action, and the selected action is determined by $\epsilon$-greedy algorithm. The exploration policy receives the same reward as the exploitation policy, and a variant of the policy gradient algorithm (REINFORCE or PPO) is used to update the policy. This approach is experimented on some typical RL problem classes, such as Pole Balancing.

Deep neural networks need a huge amount of computation for training, and this computational complexity is prohibitive sometimes. One approach to reduce the intensity of computation is through the quantization of neural networks. Basically, quantization reduces the bit-width of the operations, and it can be used to reduce the bitwidth of layers in neural networks. As accuracy preserving bitwidth may vary across different layers, the problem of learning optimal bitwidth is explored in \cite{elthakeb2019releq}. In this work, a DRL approach is proposed to determine the bitwidth of each layer. The states comprise static information about layers and dynamic information of network structure during RL training. The actions are the bitwidth of each layer, which is flexible, and the agent can change the quantization of each layer from any bitwidth to any other bitwidth. The reward pertains to the accuracy and a measure of memory and computation cost. In fact, the two objectives of the reward function are preserving accuracy and minimizing bitwidth. Using these definitions of state, action, and reward, the PPO algorithm is used to learn a policy for deriving the bitwidth of each layer of neural networks.

Components, topologies, and hyper-parameters of neural networks are automatically determined in \cite{miikkulainen2019evolving} using a neuroevolutionary algorithm in which the neural networks are trained using evolution rather than gradient descent. It can be a potential approach to determine the configuration of neural networks in deep RL algorithms. DeepNEAT starts from a population of minimal DNNs and adds nodes and edges through mutation, where each chromosome node is a network layer with its hyper-parameters, and fitness is obtained by training the decoded DNN for a fixed number of epochs. Its extension CoDeepNEAT co-evolves two populations of modules and blueprints, whose combination assembles repeated modules into a large DNN during fitness evaluation. This approach is evaluated on an image captioning task.

\textbf{Challenges.} Automating neural network architectures in deep reinforcement learning is challenging because the search is large and ill-conditioned, evaluations are noisy and sensitive to implementation details, and each candidate often requires expensive full RL training. Speedup methods such as weight sharing and proxy evaluation can reduce cost but may correlate weakly with final performance. Practical deployment further adds hardware constraints such as latency, memory, and energy, turning the problem into a multi-objective trade-off among accuracy, cost, robustness, and generalization across tasks and environments.

\section{Large Language Model for AutoRL}
\label{sec:llm}

Large Language Models (LLMs) are emerging as a practical interface between unstructured knowledge sources, including text, logs, and human intent, and reinforcement learning (RL) pipelines. From an AutoRL perspective, LLMs are particularly valuable when they reduce manual design efforts across the RL stack. Not only in reward shaping but also in algorithm and update-rule design, MDP abstraction, and policy improvement through language-level memory and feedback. In this section, we review RL-centric integrations of LLMs that directly support AutoRL components, including reward design, RL algorithm evolution, MDP automation, and LLM-augmented policy learning. Figure~\ref{fig:llmloop} illustrates the four integration paradigms.

\begin{figure*}[!t]
\centering
\resizebox{\textwidth}{!}{%
\begin{tikzpicture}[
  core/.style={draw=black!60, rounded corners=3.5pt, fill=white, align=center, font=\small, inner sep=5pt},
  role/.style={draw=cLLM!75, dashdotted, thick, rounded corners=2pt, fill=cLLM!6, align=left, font=\fontsize{8}{9.2}\selectfont, inner sep=4pt},
  tapc/.style={draw=cLLM!70!black, circle, fill=white, inner sep=1.2pt, font=\fontsize{7.5}{8}\selectfont\bfseries, text=cLLM!60!black},
  farr/.style={-{Stealth[length=2.8mm]}, very thick, black!65},
  garr/.style={-{Stealth[length=2.2mm]}, thick, cLLM!55!black},
  gd/.style={-{Stealth[length=2.2mm]}, thick, dashed, cLLM!50!black!70}]
% ---- classical loop (orthogonal, no curves) ----
\node[core, minimum width=34mm, minimum height=15mm] (agent) at (4.7,2.45) {\textbf{RL Agent}\\[-1pt]{\footnotesize policy $\pi$ $\cdot$ value function}\\[-2pt]{\footnotesize update rule}};
\node[core, minimum width=28mm, minimum height=15mm] (env) at (12.0,2.45) {\textbf{Environment}};
\draw[farr] (agent.east) -- node[above, font=\footnotesize]{action $a_t$} (env.west);
\draw[farr] (env.south) -- (12.0,0.85) -- node[below, font=\footnotesize]{$s_{t+1}$, $r_{t+1}$} (4.7,0.85) -- (agent.south);
% taps on the return segment
\node[tapc] (t2) at (5.9,0.85) {C};
\node[tapc] (t1) at (10.3,0.85) {A};
\node[tapc] (t3) at (4.7,3.2) {B};
\node[tapc] (t4) at (3.0,2.95) {D};
% ---- role modules (dash-dotted frames), straight connectors ----
\node[role, text width=46mm] (rB) at (4.7,4.85) {{\fontsize{8.5}{9.3}\selectfont\bfseries (B) LLM for algorithm evolution} {\color{black!65}(Sec.~\ref*{sec:llm-evolution})}\\ update rules, recipes, warm-started search\\ {\itshape\color{black!70}DiscoRL \cite{oh2025discorl}, OPRO \cite{yang2024opro}}};
\draw[garr] (rB.south) -- node[right, pos=0.45, font=\fontsize{7.5}{8}\selectfont, cLLM!55!black]{new update rule / recipe} (t3);
\node[role, text width=33mm] (rD) at (0.95,2.45) {{\fontsize{8.5}{9.3}\selectfont\bfseries (D) LLM as policy} {\color{black!65}(Sec.~\ref*{sec:llm-policy})}\\ memory, skill libraries, action priors\\ {\itshape\color{black!70}Voyager \cite{wang2024voyager}, Rememberer \cite{zhang2023rememberer}, ExpeL \cite{zhao2024expel}}};
\draw[garr] (rD.east|-0,2.95) -- (t4);
\node[font=\fontsize{7.5}{8}\selectfont, cLLM!55!black] at (2.9,1.08) {acts as / augments $\pi$};
\node[role, text width=40mm, anchor=north] (rC) at (5.9,-0.35) {{\fontsize{8.5}{9.3}\selectfont\bfseries (C) LLM for MDP automation} {\color{black!65}(Sec.~\ref*{sec:llm-mdp})}\\ state code, wrappers, validators\\ {\itshape\color{black!70}LESR \cite{wang2024lesr}}};
\draw[garr] (rC.north) -- node[right, pos=0.4, font=\fontsize{7.5}{8}\selectfont, cLLM!55!black]{processed state $\hat{s}$} (t2);
\node[role, text width=44mm, anchor=north] (rA) at (10.55,-0.35) {{\fontsize{8.5}{9.3}\selectfont\bfseries (A) LLM as reward designer} {\color{black!65}(Sec.~\ref*{sec:llm-reward})}\\ reward code, latent rewards, evaluate--refine\\ {\itshape\color{black!70}Eureka \cite{ma2024eureka}, LaRe \cite{qu2025latentreward}, L2R \cite{yu2023l2r}}};
\draw[garr] (10.3,-0.35) -- node[left, pos=0.4, font=\fontsize{7.5}{8}\selectfont, cLLM!55!black]{modified reward $\hat{r}$} (t1);
% ---- evaluation feedback (orthogonal, dashed) ----
\draw[gd] (env.east) -- (14.15,2.45) -- node[right, pos=0.5, font=\fontsize{7.5}{8}\selectfont, black!70, align=left, rotate=90, anchor=north]{evaluation feedback} (14.15,-1.0) -- (12.85,-1.0);
\end{tikzpicture}}
\caption{LLM-enhanced AutoRL: the four LLM roles of Section~\ref{sec:llm} (A--D) and the loop edge each one intercepts; the dashed arrow closes the evaluate--refine loop.}
\label{fig:llmloop}
\end{figure*}

\subsection{LLM for Reward Design}
\label{sec:llm-reward}

Reward design remains a central bottleneck in RL, particularly in scenarios involving sparse feedback and long-horizon credit assignment. Recent work indicates that LLMs can contribute several RL-grounded capabilities to automate this process. First, LLMs can translate natural-language objectives into structured reward templates or executable reward code. For example, Yu et al.~\cite{yu2023l2r} map language instructions to reward specifications for robotic skill synthesis, providing a practical route to turn high-level intent into trainable reward signals. Building on this, LLMs can improve credit assignment by converting episodic feedback into more informative intermediate signals. Qu et al.~\cite{qu2025latentreward} propose \textsc{LaRe}, which leverages LLMs to restructure episodic rewards into intermediate latent rewards, thereby enhancing credit assignment and bridging the gap between linguistic priors and symbolic reward requirements.

Furthermore, LLMs can generate and iteratively revise reward code through an evaluate--refine loop. Ma et al.~\cite{ma2024eureka} demonstrate that LLMs can produce reward programs for continuous-control tasks, with iterative refinement yielding reward functions that outperform expert-designed baselines across diverse robotic environments. Quantitatively, Eureka outperforms human-expert rewards on $83\%$ of $29$ open-source RL environments, with an average normalized improvement of $52\%$ \cite{ma2024eureka}. To extend these capabilities, LLM-based reward design can be strengthened by incorporating richer feedback modalities (e.g., demonstrations and vision-language cues) to mitigate the limitations of text-only specifications. Chen et al.~\cite{chen2025elemental} introduce an interactive framework that combines demonstrations with language/vision-language modeling to align reward features and weights more reliably for robotics.

Finally, beyond shaping and templating, reward functions can be discovered as an optimization objective. Lu et al.~\cite{lu2025rewarddiscovery} introduce a bilevel framework that searches for reward functions for embodied RL agents via regret minimization, thereby reducing dependence on handcrafted rewards. These advancements in reward automation set the stage for broader LLM applications in RL algorithm evolution.

\subsection{LLM for RL Algorithm Evolution}
\label{sec:llm-evolution}

Building on the automation of reward functions, a core objective of AutoRL is to automate the learning process itself, extending beyond \emph{what} is learned to \emph{how} learning occurs. LLMs can facilitate RL algorithm evolution at two levels: discovering update rules and generating training recipes or configurations for outer-loop optimizers. A significant advancement in automated RL algorithm design involves searching directly in the space of learning rules. \cite{oh2025discorl} shows that machines can discover state-of-the-art RL update rules that surpass manually designed counterparts across challenging benchmarks, establishing a foundation for algorithm evolution beyond hyper-parameter tuning. The discovered rule reaches an interquartile-mean score of $13.86$ on Atari-57, surpassing MuZero, and outperforms published methods on the held-out ProcGen benchmark without any modification \cite{oh2025discorl}. In addition, even with a fixed base algorithm family, AutoRL requires effective selections of learning-rate schedules, regularization, normalization, replay/batching details, and other recipe-level elements. This need is reinforced by empirical evidence that RL performance can be highly sensitive to hyper-parameter choices and tuning protocols. \cite{eimer2023hprl} systematically analyzes this sensitivity and recommend principled HPO practices for reproducible RL. Within such limited-budget search scenarios, LLMs can serve as proposal models that shape the search space and warm-start outer-loop optimizers. \cite{yang2024opro} explores LLMs as optimizers through prompting, while \cite{zhang2023llmhpo} studies LLM-guided decisions in hyper-parameter optimization under constrained evaluation budgets. In AutoRL, these LLM proposals can restrict the search to plausible regions and propose better initial configurations, while final selections remain grounded in empirical evaluation.

Overall, this sub-area reframes AutoRL as a human-knowledge-conditioned search over RL algorithms and recipes, where LLMs provide priors and structure, and the AutoRL loop ensures ground-truth validation. This   flexibility complements LLM-driven efforts in automating MDP components.

\subsection{LLM for MDP Automation}
\label{sec:llm-mdp}

Complementing the evolution of RL algorithms, LLMs can automate the construction of MDP components by translating raw problem contexts into RL related abstractions. In complex domains with structured logs or mixed discrete-continuous contexts, LLMs can summarize global system states into representations more suitable for RL. Wang et al.~\cite{wang2024lesr} propose \textsc{LESR}, where an LLM generates task-relevant state representation code that incorporates domain priors, thereby improving sample efficiency and downstream policy learning. LESR improves accumulated reward by $29\%$ on MuJoCo tasks and success rate by $30\%$ on Gym-Robotics compared with state-of-the-art baselines \cite{wang2024lesr}. Moreover, many real-world RL problems demand adherence to validity constraints and operational rules. LLMs can express these as validators or action schemas (e.g., parameterized actions or hierarchical options), enabling AutoRL to search over action abstractions while enforcing feasibility at execution time. Additionally, when interactions involve APIs, simulators, or tool protocols, LLMs can generate wrappers to normalize reset, step, termination, and observation formatting, enhancing the portability of AutoRL pipelines across domains. These MDP enhancements pave the way for LLMs to serve directly as policy learners.

\subsection{LLM as Policy Learner}
\label{sec:llm-policy}

Extending from MDP automation, LLMs can function as policy components, often by integrating language-level planning and memory with RL-style improvement mechanisms. A foundational direction is to use pre-trained language models as general-purpose decision-making backbones, adapting them to interactive environments. Li et al.~\cite{li2022plmidm} study how pre-trained language models can be used for interactive decision-making, providing an early blueprint for language-model-based policies in sequential environments. Building on this, Zhang et al.~\cite{zhang2023rememberer} introduce \textsc{Rememberer}, which equips an LLM with long-term experience memory and an RL mechanism for updating it, enabling continual improvement without fine-tuning model parameters. Zhao et al.~\cite{zhao2024expel} present \textsc{ExpeL}, where an agent collects experiences, extracts reusable natural-language insights, and applies them as self-generated guidance during inference. Wang et al.~\cite{wang2024voyager} propose \textsc{Voyager}, which builds an expanding library of executable skills through an iterative feedback loop, facilitating long-horizon behavior via compositional skill reuse. 
Moreover, language priors can be injected into RL more explicitly to improve sample efficiency. Yan et al.~\cite{yan2025llmprior} treat LLM outputs as action priors and integrate them into RL through Bayesian-style inference, showing that prior knowledge from LLMs can reduce exploration burden and accelerate learning. At the intersection of reward and policy learning, Du et al.~\cite{du2023ellm} show that LLM-suggested goals can guide exploration and pretraining for RL agents, injecting language priors into the process without relying on dense hand-crafted rewards.

\subsection{Challenges in LLM Integration}
While LLMs provide powerful tools for automating reward design, RL algorithms, MDP construction, and policy learning, their integration into AutoRL pipelines introduces several reliability challenges. Generated elements—such as reward functions, algorithm configurations, MDP abstractions, or policy guidance—may suffer from inconsistencies, incompleteness, or overfitting to specific benchmarks, potentially undermining reproducibility and generalizability across diverse tasks. The outer-loop optimization process becomes particularly vulnerable to variations in prompt engineering and context retrieval, where minor alterations in instructions or supporting evidence can lead to divergent algorithmic choices or suboptimal proposals. Moreover, seamlessly incorporating LLM suggestions into RL workflows demands rigorous verification mechanisms, safety protocols, and resource management strategies, as erroneous outputs could squander computational budgets, yield policies that exploit unintended reward loopholes, or breach operational constraints, thereby offsetting the efficiency gains promised by automation.
 
Overall, the results in this section and Section~\ref{sec:hyp} show that 
automating different RL components consistently results in double-digit improvements over expert-designed baselines, highlighting the value of the AutoRL outer loop. Limitations in comparing results across studies are discussed in Section~\ref{sec:limit}.

\section{Limitations and Future Work}
\label{sec:limit}
Existing AutoRL approaches face several limitations, as highlighted in the literature. A primary limitation is the lack of comparable and reproducible evaluation. Deep RL results are highly sensitive to evaluation protocols which makes fair cross-paper comparisons difficult~\cite{henderson2018deep}. This variability also increases the risk that outer-loop search overfits to a specific benchmark setup rather than producing robust improvements~\cite{henderson2018deep,eimer2023hprl}. Current AutoRL benchmarks are often limited to simulation suites like OpenAI Gym or DeepMind Control Suite, which feature low-dimensional, dense-reward tasks. Methods tuned on these benchmarks may struggle to generalize to complex, sparse-reward, or partially observable environments. AutoRL can be computationally demanding because it typically requires many full training runs, and the cost grows quickly with longer horizons, high-dimensional observations, or complex action spaces~\cite{eimer2023hprl}. Therefore, the scalability is a severe challenge for high-dimensional tasks or long-horizon problems. 

On the other hand, AutoRL introduces an additional optimization layer on top of standard RL, which increases computational cost. In a conventional RL setup, the training complexity depends on the number of training episodes, the episode length, and the cost of forward and backward passes through the policy and value networks. 
% Let the cost of training a single RL agent be denoted as $C_{\mathrm{RL}}$. AutoRL evaluates multiple candidate configurations (e.g., hyper-parameters, architectures, reward functions). 
If $K$ configurations are explored, in the simplest case, where all candidates are fully trained, the overall computational cost scales approximately linearly with $K$.
% i.e.,
% \begin{equation}
% C_{\mathrm{AutoRL}} \approx K \cdot C_{\mathrm{RL}} .
% \end{equation}
The dominant cost typically arises from repeated RL training. More efficient strategies, such as Bayesian optimization, evolutionary methods with parallel evaluation, and early stopping, can significantly reduce practical overhead and wall-clock time. Although AutoRL increases total computational requirements, it can improve robustness and reduce repeated manual tuning, making the additional cost justifiable in many complex applications.

Moreover, RL performance can vary substantially across random seeds and minor implementation choices, so AutoRL may select configurations that appear strong largely by chance without rigorous multi-seed controls and statistically meaningful reporting~\cite{henderson2018deep}. Finally, many AutoRL methods are validated mainly in simulated settings and may overlook real-world constraints such as safety, feasibility, delayed feedback, and operational costs, which are central in deployment~\cite{achiam2017cpo,ray2019safetygym}. In addition, automation layers introduced for representation or training can bring additional hyper-parameters, which may shift manual effort rather than eliminate it if they are not automated end-to-end~\cite{afshar2020state,ivanovic2019barc}.

These limitations suggest that progress in AutoRL should be measured not only by peak performance, but also by comparability, efficiency, robustness, and deployment readiness. A first direction is standardized AutoRL benchmarks and protocols that target the automation problem itself. Such benchmarks should fix compute budgets, require multi-seed reporting with uncertainty, and emphasize generalization so that different AutoRL methods can be compared under matched resources~\cite{henderson2018deep,cobbe2020procgen,gulcehre2020rlunplugged}; ARLBench \cite{becktepe2024arlbench} is a first step in this direction. Such protocols are particularly important because the headline results reported by state-of-the-art AutoRL methods, including those quoted in Sections \ref{sec:hyp} and \ref{sec:llm}, each follow their own budgets, seeds, and evaluation conventions and are therefore not directly comparable across papers. A second direction is budget-aware scalability. Practical AutoRL should reduce the number of full training runs via multi-fidelity optimization, principled early stopping, and reuse of information across trials, which is especially important when search spaces become large~\cite{li2017hyperband,eimer2023hprl}. A third direction is robustness under noise: AutoRL objectives and selection rules should account for variance to avoid randomly good selection and improve reproducibility~\cite{henderson2018deep,eimer2023hprl}. Beyond these, real-world deployment calls for constraint-aware AutoRL pipelines that integrate safety and feasibility directly into MDP modeling and configuration search, building on safe RL formulations and benchmarks~\cite{achiam2017cpo,ray2019safetygym}. Deployment also requires bridging simulation-to-reality gaps in embodied settings, where domain shift motivates robustness-oriented evaluation and techniques such as domain randomization~\cite{tobin2017domain}. Finally, LLM-assisted AutoRL is a promising direction for reducing manual design effort by generating candidate reward templates, abstractions, and training recipes as \emph{proposals} that are then validated within the AutoRL outer loop~\cite{ma2024eureka}. Since LLM outputs can be inconsistent or biased, future systems should include verification mechanisms such as constraint checks and self-validation, and rely on empirical evaluation to prevent optimizing spurious signals~\cite{qu2025latentreward,ma2024eureka}.

\section{Ethical Considerations and Potential Risks}
\label{sec:ethics}

AutoRL can reduce the need for expert intervention and speed up RL development. However, greater automation also raises important ethical and practical concerns that should not be overlooked. One central issue is reward misspecification. When reward functions are generated or tuned automatically, the system may optimize for signals that improperly reflect the true objective. As a result, the agent can learn behaviors that technically maximize reward but contradict the intended goal. In safety-critical domains such as robotics, healthcare, or autonomous systems, such misalignment may lead to unsafe or undesirable outcomes.
Another concern relates to bias in learned representations. Automated state abstraction methods determine which features of the environment are emphasized or ignored. If the training data or optimization process contains biases, these may be embedded in the learned representation, potentially leading to unfair or systematically skewed decisions. Automation may also encourage unsafe exploration: without appropriate constraints, aggressive exploration by evolutionary or curiosity-driven methods can produce harmful actions during training, and growing automation reduces human oversight. Automatically generated reward functions or representations can make system behavior harder to interpret and audit. This reduced transparency can complicate accountability, especially in regulated or high-stakes applications.

For these reasons, automating RL should be accompanied by safeguards such as human-in-the-loop validation, safety constraints, fairness assessments, and systematic testing under diverse scenarios.

\section{Conclusion}
\label{sec:conc}

In this paper, we have presented recent work in automated reinforcement learning that can be incorporated into an automated RL or DRL pipeline. We also introduce a general AutoRL pipeline suitable for solving sequential decision-making problems. This field is gaining increasing popularity, as a robust and high-quality reinforcement learning pipeline can address many complex tasks while significantly reducing required time and resources. The RL framework is divided into three main components in this paper, with relevant work discussed for each: MDP modeling, algorithm selection, and hyper-parameter optimization. In addition, learning-to-learn methods and neural network architecture automation are covered in separate sections. Through our exploration of the AutoRL literature, we conclude that a concrete and complete pipeline for AutoRL, analogous to those in AutoML, has yet to be fully developed, despite its substantial benefits for designing and solving sequential decision-making problems. Key research questions include optimizing hyper-parameters with minimal resources, automatically modeling problems as MDPs, and generalizing mappings from available information to RL environments. The integration of large language models into AutoRL pipelines represents a promising direction, as evidenced by recent advances in LLM-assisted reward design, algorithm evolution, and policy learning.

\bibliographystyle{IEEEtran}
\bibliography{sn-bibliography-r3}

\end{document}